\documentclass{article}

\usepackage[utf8]{inputenc} % allow utf-8 input
\usepackage[T1]{fontenc}    % use 8-bit T1 fonts
\usepackage{url}            % simple URL typesetting
\usepackage{booktabs}       % professional-quality tables
\usepackage{amsfonts}       % blackboard math symbols
\usepackage{nicefrac}       % compact symbols for 1/2, etc.
\usepackage{microtype}      % microtypography

\usepackage{graphicx}
\usepackage{subcaption}
\usepackage{makecell,multirow} %

\usepackage{xcolor}
\definecolor{niceblue}{rgb}{0.0,0.19,0.56}
\usepackage[breaklinks]{hyperref}

\usepackage{breakurl}
\hypersetup{colorlinks,linkcolor={blue},citecolor={niceblue},urlcolor={blue}}
\usepackage{amsmath}

\usepackage{caption}
\usepackage{mwe}

\usepackage{algorithm}
\usepackage{algorithmic}
\usepackage[final,nonatbib]{neurips_2021}
\usepackage{xcolor}
\usepackage{enumitem}
\usepackage{optidef}
\usepackage{soul}
\usepackage{textcomp}  % Required for encoding \textbigcircle
\usepackage{scalerel}  % Required for emoji \scalerel
\sethlcolor{lightgray}
\def\hourglass{\scalerel*{\includegraphics{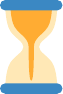}}{\textrm{\textbigcircle}}}
\def\checkbox{\scalerel*{\includegraphics{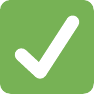}}{\textrm{\textbigcircle}}}
\DeclareSymbolFont{matha}{OML}{txmi}{m}{it}% txfonts
\DeclareMathSymbol{\varv}{\mathord}{matha}{118}

\newcommand{\block}[1]{%
  \raisebox{\dimexpr(\fontcharht\font`X-1em)/2}{\rule{1em}{#1\dimexpr1em/8}}%
}
\DeclareUnicodeCharacter{2581}{\block{1}}

\title{Distributed Deep Learning In Open Collaborations}

\author{%
  Michael Diskin\thanks{Equal contribution. Correspondence to \texttt{mryabinin0@gmail.com}\newline Detailed author contributions are listed \hyperref[sec:contributions]{at the end} of the work.}\textsuperscript{\hspace{0.6em}$\dag\heartsuit$}
   \And
   Alexey Bukhtiyarov\footnotemark[1]\textsuperscript{\hspace{0.6em}$\dag\clubsuit$}
   \And
   Max Ryabinin\footnotemark[1]\textsuperscript{\hspace{0.6em}$\dag\heartsuit$}
   \AND
   Lucile Saulnier\textsuperscript{$\ddag$}
   \And
   Quentin Lhoest\textsuperscript{$\ddag$}
   \And
   Anton Sinitsin\textsuperscript{$\dag\heartsuit$}
   \And
   Dmitry Popov\textsuperscript{$\dag\heartsuit$}\\
   \And
   Dmitry Pyrkin\textsuperscript{$\heartsuit$}\\
   \And 
   Maxim Kashirin\textsuperscript{$\heartsuit$}\\
   \And
   Alexander Borzunov\textsuperscript{$\dag\heartsuit$}\\
   \And
   Albert Villanova del Moral\textsuperscript{$\ddag$}\\
   \And
   Denis Mazur\textsuperscript{$\clubsuit$}\\
   \And
   Ilia Kobelev\textsuperscript{$\dag\clubsuit$}\\
   \And
   Yacine Jernite\textsuperscript{$\ddag$}\\
   \And
   Thomas Wolf\textsuperscript{$\ddag$}\\
   \And
   Gennady Pekhimenko\textsuperscript{$\diamondsuit\spadesuit$}\\
   \and
   \centerline{$\dag$ Yandex, Russia}\\
   \centerline{$\ddag$ Hugging Face, USA}\\
   \centerline{$\heartsuit$ HSE University, Russia}\\
   \centerline{$\clubsuit$ Moscow Institute of Physics and Technology, Russia}\\
   \centerline{$\diamondsuit$  University of Toronto, Canada}\\
   \centerline{$\spadesuit$ Vector Institute, Canada}
\vspace{-16pt}
}

\begin{document}

\maketitle

\begin{abstract}
Modern deep learning applications require increasingly more compute to train state-of-the-art models. To address this demand, large corporations and institutions use dedicated High-Performance Computing clusters, whose construction and maintenance are both environmentally costly and well beyond the budget of most organizations. As a result, some research directions become the exclusive domain of a few large industrial and even fewer academic actors. To alleviate this disparity, smaller groups may pool their computational resources and run collaborative experiments that benefit all participants. This paradigm, known as grid- or volunteer computing, has seen successful applications in numerous scientific areas. However, using this approach for machine learning is difficult due to high latency, asymmetric bandwidth, and several challenges unique to volunteer computing. In this work, we carefully analyze these constraints and propose a novel algorithmic framework designed specifically for collaborative training. We demonstrate the effectiveness of our approach for SwAV and ALBERT pretraining in realistic conditions and achieve performance comparable to traditional setups at a fraction of the cost. Finally, we provide a detailed report of successful collaborative language model pretraining with 40 participants.
\vspace{-8pt}

\end{abstract}
\section{Introduction}\label{sect:intro}

The deep learning community is becoming increasingly more reliant on transfer learning. In computer vision, pretraining convolutional networks on large image collections such as ImageNet~\cite{imagenet_cvpr09} is the de facto standard for a wide range of applications ranging from object detection~\cite{10.1109/CVPR.2014.81} and semantic segmentation~\cite{7298965} to image classification~\cite{pmlr-v32-donahue14} and even learning perceptual similarity~\cite{johnson2016perceptual}. A growing number of natural language processing systems capitalize on language models with billions of parameters~\cite{bert,albert,roberta,xlmr,gpt3,shoeybi2019megatron} trained on vast unlabeled corpora. Similar trends have emerged in areas such as speech processing~\cite{baevski2020wav2vec}, reinforcement learning~\cite{zhu2021transfer}, and computational biology~\cite{Lu2020.09.04.283929,honda2019smiles}.

Training these models is a notoriously time-consuming and challenging task: it often requires hundreds of high-end GPU servers~\cite{gpt3,megatron2} and would take multiple years on a single device~\cite{lin2020multinode}. Most academic and independent researchers simply cannot afford to train state-of-the-art models from scratch, which slows down scientific progress and practical adoption of deep learning.

Historically, the deep learning community has addressed this problem via ``model hubs'' or ``model zoos'' --- public repositories for pretrained model checkpoints~\cite{tfhub,torchhub,hfhub,dlhub}. These repositories have played a significant role in the democratization of deep learning, allowing everyone to reap the benefits of large-scale training runs conducted by corporations and universities with sufficient resources. However, model hubs are limited to a narrow subset of datasets and tasks that match the interests of model creators. For instance, in natural language processing, it is often difficult to find up-to-date models for more than a handful of languages~\cite{Joshi2020TheSA}. In turn, computer vision hubs rarely feature models trained on drawings, satellite images, 3D renders, microscopy, or any other data that does not resemble ImageNet. As a result, many researchers in these areas can only work on problems for which there are available pretrained models rather than the problems that most need solving.

However, there might be an alternative way to obtain pretrained models: to train these models \emph{collaboratively}. This approach, known as volunteer (or grid) computing, allows many independent parties to combine their computational resources and collectively perform large-scale experiments~\cite{seti_at_home,foldingathome,anderson2004boinc}. The raw compute performance of such collaborations often exceeds that of the fastest supercomputers~\cite{folding_exaflop_2}; however, fully utilizing it can be challenging due to several reasons. First, devices that contribute to collaborative experiments can range from GPU servers and high-end workstations to consumer-grade computers and even smartphones~\cite{Tapparello2016VolunteerCO}. Second, most of these devices use household internet connection with limited bandwidth and low reliability. Third, participants in such projects often donate their hardware part-time, joining and leaving the experiment at will.

While it is theoretically possible to train neural networks on this kind of infrastructure, modern distributed training strategies are only efficient in a narrow range of conditions. For instance, training with Ring All-Reduce~\cite{ringallreduce} works well for identical servers but suffers significant performance penalties from network latency or bandwidth variation~\cite{wagma}. Another technique known as Parameter Server can handle heterogeneous devices at the cost of being less scalable~\cite{ps}. Applying any of these strategies outside their preferred conditions may significantly reduce the training throughput~\cite{shi2018performance}, which makes them difficult to apply in the volatile infrastructure of volunteer computing. This issue is further complicated by the unique limitations of volunteer devices, such as network address translation (NAT), regional access restrictions, or variations in performance.

In this study, we carefully analyze the above challenges and come up with a practical solution for \textbf{D}istribut\textbf{e}d \textbf{D}eep \textbf{L}earning in \textbf{O}pen \textbf{C}ollaborations (DeDLOC). DeDLOC is based on a novel algorithm that adapts to the available hardware in order to maximize the training throughput. Depending on the infrastructure, DeDLOC can recover parameter servers~\cite{ps}, All-Reduce SGD~\cite{sergeev2018horovod}, decentralized SGD~\cite{dp_sgd}, BytePS~\cite{byteps}, or an intermediate strategy that combines all of them. Using this algorithm, we propose a system for collaborative training designed to accommodate a large number of heterogeneous devices with uneven compute, bandwidth, reliability, and network capabilities.

The contributions of our work can be summarized as follows:

\begin{itemize}[leftmargin=*]
    \item We analyze the unique challenges of distributed training in open collaborations and propose a practical recipe for training in these conditions.
    \item We formulate a novel distributed training algorithm that interpolates between traditional strategies to directly maximize the training performance for the available hardware.
    \item We verify the effectiveness of the proposed algorithm and system design for unsupervised pretraining of ALBERT-Large and SwAV under realistic conditions.
    \item We run collaborative training with actual volunteers, achieving competitive results to models trained on hundreds of data center GPUs. We also report insights on the collaborator activity and share the codebase for running similar experiments in the future\footnote{Code and training configurations are available at \href{https://github.com/yandex-research/DeDLOC}{\texttt{github.com/yandex-research/DeDLOC}}}.
\end{itemize}

\section{Related work}\label{sect:related}

\subsection{Distributed training}\label{sect:related_distributed}

In this work, we focus on distributed data-parallel training, where each device runs forward and backward pass of the entire model on a subset of training examples. While there are many alternative techniques~\cite{huang2019gpipe,shazeer2017outrageously,zero}, data-parallel is still the most popular strategy. Even the model-parallel approaches for extremely large models rely on data parallelism at the top level~\cite{zero,megatron2,switch}.

Training on multiple nodes was first implemented with parameter server (PS)~\cite{ps}. This training strategy relies on a dedicated node that stores model parameters and executes optimization steps using the gradients sent by workers. 
In turn, worker nodes iteratively download the latest version of model parameters from the server, compute gradients and submit them back to the PS. This strategy is easy to implement and use, but it has an unavoidable bottleneck: the entire system performance is limited by the network throughput of a single server. Since then, the scientific community proposed numerous extensions to PS that alleviate the bottleneck by reducing the communication load~\cite{deepgradientcompression,localsgd_first,Stich18local,koloskova2020decentralized,li2020acceleration}, introducing asynchronous updates~\cite{recht2011hogwild,projectadam} or training with multiple servers~\cite{sharded_ps_first, byteps}.

% A recent extension of this method, called BytePS~\cite{byteps}, is particularly relevant in the context of our work: its authors propose to use several CPU-only parameter servers to balance the network load across the nodes. The adaptive algorithm proposed in our work can obtain this architecture as a special case given the communication capabilities of each participant; we elaborate on this result in Section~\ref{sect:method_algorithm}. 

The issue of uneven communication load has also inspired the development and widespread adoption of another group of methods that rely on All-Reduce for gradient averaging~\cite{goyal2017accurate,mikami2019massively,lamb}. All-Reduce is a family of collective operations that allow nodes to efficiently aggregate (e.g. average) their local vectors and distribute the result across all devices~\cite{ringallreduce,mpich_rabenseifner,torus_allreduce}. Unlike parameter servers, All-Reduce assigns equal roles to all devices, making it easier to scale to a large number of homogeneous workers.

The popularity of AR-SGD sparked many practical applications for different scenarios. One particularly relevant application is elastic training~\cite{pytorch_elastic,elastic_horovod}, which allows the user to add or remove workers at any point without interrupting the training run.
While this bears a lot of similarity with collaborative training, we have found that elastic training systems are designed around global state synchronization, which makes them highly dependent on the homogeneity of the workers and their network connectivity. The overall efficiency is bounded by the performance of the lowest-performing node; as a result, introducing even a single low-bandwidth participant to such systems reduces the training speed by orders of magnitude.

Seeking to avoid the need for synchronization and centralized orchestration, the research community has developed decentralized training algorithms. These algorithms can be broadly divided into two categories: directly passing updates between peers~\cite{sgp,slowmo} or running All-Reduce in small alternating groups~\cite{moshpit,wagma}. Compared to PS and All-Reduce, both categories provide a greater degree of fault tolerance but often require more steps to converge due to delayed updates~\cite{dp_sgd,wagma}.

Most practical use cases of the above techniques take place in HPC or cloud conditions, but there is one notable exception. In Federated Learning, multiple parties train a shared model on decentralized privacy-sensitive data that cannot be shared between devices~\cite{FedLearningOriginal}. For that reason, federated learning algorithms prioritize data privacy over training efficiency, often leaving most of the compute resources unused~\cite{FedLearningAtScale,FedLearningDecentralized}. For a more detailed overview of Federated Learning, refer to Appendix~\ref{appendix:related_federated}.

\subsection{Volunteer Computing}\label{sect:related_volunteer}

Volunteer computing (VC) is a paradigm of distributed computing where people donate the idle time of their desktops, smartphones, and other personal devices to solve a computationally hard problem collectively. This approach has seen successful applications in bioinformatics, physics and other scientific areas~\cite{larson_crowd, folding_covid, lhc_at_home, seti_at_home, qmc_at_home, folding_timeline,einstein_at_home}.

In all these applications, volunteer computing allows researchers to access vast computational resources. In Folding@home, over 700,000 volunteers have collectively contributed 2.43 exaFLOPs of compute to COVID-19 research in April of 2020~\cite{folding_exaflop_2}. Another project named BOINC (Berkeley Open Infrastructure for Network Computing) brings together 41.548 petaFLOPs from over 790,000 active computers as of 17 March 2020~\cite{anderson2004boinc}. Volunteer computing systems were also the first ``supercomputers'' to reach 1 petaFLOP and 1 exaFLOP barriers~\cite{folding_exaflop_2, folding_petaflop}. These results became possible due to the contributions of a broad range of devices from high-end workstations to smartphones and even gaming consoles~\cite{folding_ps3}.

Unfortunately, this compute diversity is also the main limitation of VC. Any volunteer computing system should be able to run on a wide range of available hardware and maintain integrity even if some participants disconnect. Furthermore, the resources available to a project can vary over time, as most volunteers are only sharing their hardware when it is unused. Finally, volunteer devices are interconnected with a shared high latency network at typical home internet connection speeds.

As a result, there were only a few successful attempts to apply volunteer computing to machine learning workloads. One such project is MLC@Home~\cite{clemens2021mlds}, which relies on volunteers to train many small independent models. 
This specific problem can be solved with no direct communication between participants. By contrast, distributed training of a single model requires significantly more communication and does not allow a natural way to ``restart'' failed jobs. When it comes to distributed training of neural networks, most volunteer computing projects rely on parameter server architectures~\cite{lc0,volunteer_dl_async,atre2021distributed}. As a result, these systems are bounded by the throughput of parameter servers and the memory available on the weakest GPU. The only notable exception is Learning@home~\cite{hivemind_dmoe}, which uses expert parallelism to train larger models spanning multiple computers; however, this approach has only been tested in simulated conditions.

\section{Distributed Deep Learning in Open Collaborations}
\label{sect:method}

There are two unsolved challenges that stand in the way of practical collaborative training. The first challenge is algorithmic: how to maintain optimal training performance with dynamically changing hardware and network conditions? Another major challenge is ensuring consistent training outcomes with inconsistent composition of participants. 
Thus, we organize this section around these two issues:

\begin{itemize}[leftmargin=*]
    \item Section~\ref{sect:method_general} provides a general overview of DeDLOC and explains how it maintains consistency in a dynamic environment.
    \item In Section~\ref{sect:method_algorithm}, we describe the generalized communication strategy that maximizes training throughput by adapting to the currently available devices.
    \item In Section~\ref{sect:method_system_design}, we address system design challenges, such as circumventing NAT and firewalls, training on large datasets and managing collaborator access.
\end{itemize}

\subsection{Ensuring training consistency}\label{sect:method_general}

Many state-of-the-art models, notably GANs~\cite{GAN} and Transformers~\cite{transformer}, require a strict training regimen. Deviating from the recommended batch size or introducing stale gradients may significantly affect the training outcome~\cite{trainingtips,liu-etal-2020-understanding,pipedream}. Since in a collaborative setting one has little control over the devices that participate in the experiment, it is almost guaranteed that the specific hardware setup will vary between runs and even during a single run. Without special precautions, these runs may result in models with vastly different final accuracy.

\begin{figure}[b]
\vspace{-10pt}
    \centering
    \includegraphics[height=100px]{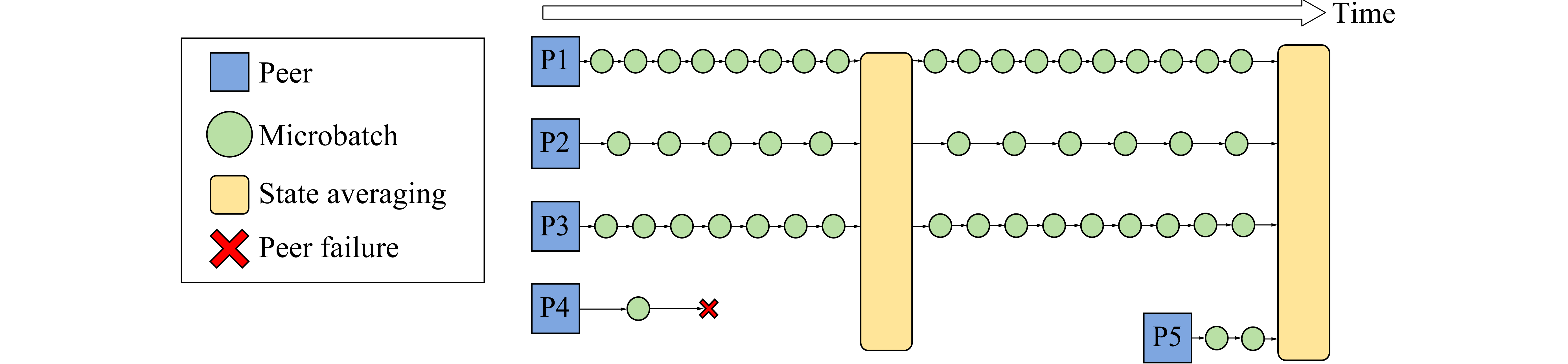}
    \vspace{-8pt}
    \caption{Two DeDLOC training iterations with example collaborator dynamics.}
    \label{fig:collaborative_training}
    \vspace{-12px}
\end{figure}

To avoid this pitfall, DeDLOC follows synchronous data-parallel training with fixed hyperparameters regardless of the number of collaborators. In order to compensate for relatively slow communication, we adopt training with extremely large batches~\cite{lars,lamb,trainingtips,kaplan2020scaling,gpt3}, which allows peers to communicate less frequently. This strategy also provides a natural way to deal with heterogeneous hardware~\cite{ott2018scaling}: each device accumulates gradients at its own pace until the collaboration reaches the target batch size. Once ready, the collaborators exchange their gradients and perform one optimizer step. Using synchronous updates makes DeDLOC mathematically equivalent to large-batch training on a regular HPC cluster; see Appendix~\ref{appendix:convergence_analysis} for a more detailed explanation. Figure~\ref{fig:collaborative_training} gives a high-level visual explanation of this algorithm.

\subsection{Adaptive averaging algorithm}\label{sect:method_algorithm}

As we discussed in Section~\ref{sect:related_distributed}, each distributed training algorithm has a narrow range of conditions where it can reach optimal performance. For instance, Ring All-Reduce works best on homogeneous hardware with low-latency communication, while Parameter Server strategy requires dedicated high-bandwidth devices that communicate with a large number of ``workers''. Since all devices are provided by volunteers, our training infrastructure is in a constant state of flux.

For instance, a collaboration can start with several homogeneous nodes that could be trained optimally with All-Reduce. If new participants bring devices with less bandwidth, it may be more efficient to use the original nodes as parameter servers. As more peers join, these servers will eventually become unable to handle the network load and the collaboration will need to switch to a different strategy.

Running efficient training on this kind of infrastructure requires a protocol that can dynamically assign roles to every peer given their hardware and network capabilities:
\begin{itemize}[leftmargin=*]
    \item \textbf{Compute performance:} Each peer $i \in 1,\ \dots,\ n$ can compute gradients over $s_i$ samples per second. A peer that is unable to compute gradients (i.e. that has no GPU) will have $s_i{=}0$.
    \item \textbf{Bandwidth:} Peers communicate with a limited throughput: $d_i$ for download and $u_i$ for upload.
    \item \textbf{Geographical limitations:} In addition to individual bandwidth, the communication throughput between two peers $i, j$ is also restricted by $t_{i j}$ and $t_{j i}$ in each direction.
\end{itemize}

Given these constraints, our objective is to find a communication strategy that has the highest training throughput, that is, the one that \textit{makes the most SGD steps with a target batch size $B$ per unit of time}. In turn, the training throughput of a collaboration depends on how we split the load among the participants. Each peer can be assigned to compute gradients over a subset of training examples, aggregate a part of those gradients from all peers, or both.

For simplicity and efficiency, we use delayed parameter updates (DPU)~\cite{zerooffload} --- a technique that allows gradient computation and communication to run in parallel, at the cost of exactly one round of staleness. This strategy can improve time to convergence for a wide range of models, including Transformers~\cite{zerooffload,aji2019making}. That said, our approach can be easily adapted to non-concurrent updates.

With DPU, the frequency of training updates is determined by either the time to compute gradients or the time to aggregate them, whichever takes longer. In total, a collaboration processes $\sum_{i=1}^{n} s_i \cdot c_i$ samples per second, where $c_i$ is the binary indicator denoting whether $i$-th peer is assigned to contribute gradients. Assuming the target batch size $B$, the frequency of the computation phase can be expressed as $F_{compute} = \sum_{i=1}^{n} s_i \cdot c_i \text{  } / \space B$.

During the communication phase, each peer is first assigned to accumulate gradients over a fraction of model parameters. After that, everyone partitions their local gradients and sends each partition to the corresponding peer. On the other end, receiver nodes accumulate the gradients from all senders and return the average. In modern distributed training systems, this procedure is highly parallelized~\cite{byteps,pytorch_distributed}: a reducer can aggregate one chunk of gradients while downloading the next chunk and distributing the previous one back to the same senders.

In order to properly optimize the training throughput, we must account for this parallelism. As such, we explicitly define the speed $a_{i j}$ at which peer $i$ sends gradients to peer $j$ for aggregation. In turn, $j$-th peer aggregates gradients from all peers at the rate of the slowest sender $a_j = \min_{i : c_i = 1} a_{i j}$. The senders can then get the aggregated results from the $j$-th reducer at $g_{j i} \leq a_j$. Finally, the total $a_{i j}$ and $g_{i j}$ for each peer cannot exceed their maximum download/upload speed. The only exception is that transfer within one node ($a_{i i}, \ g_{i i}$) does not count towards network throughput.

The frequency of the gradient aggregation phase is simply the rate at which the slowest peer can aggregate the full gradient vector: $F_{agg} = \min_i {\sum_{j} g_{j i}} \text{ } / \text{ }P$ , where $P$ is the number of model parameters. The final optimization problem can be formulated as follows:
% TODO (mryab) \include{optidef}
\begin{equation}
\label{eq:main_problem}
\begin{array}{rclll}
\underset{a, g, c}{\max} & &
                                \min\Bigg(\frac{\sum_{i=1}^n s_i \cdot c_i }{B},\ 
                                     \frac{ \min_i\sum_{j} g_{j i}}{P}\Bigg)&\quad&\\
\textrm{s.t. } &\quad & g_{i j} \leq \min_{k: c_k{=}1} a_{k i}  &\quad&\forall i, j \\
                 & \quad  & \sum_{j \neq i}\left( a_{j i} + g_{j i}\right) \leq d_{i} & \quad&\forall i \\
                 & \quad &   \sum_{j \neq i}\left( a_{i j} + g_{i j}\right) \leq u_{i} & \quad&\forall i \\
                 & \quad &   a_{i j} + g_{i j} \leq t_{i j} & \quad&\forall i, j \\
\end{array}
\end{equation}

This problem must be solved regularly as participants are joining and leaving. Thus, we must ensure that the benefits of the optimal strategy outweigh the overhead of computing it. For that reason, we formulate optimal strategy search as a linear program that can be solved efficiently\footnote{In our experiments, the LP solver consistently converged in  $<$ 50ms and was called $\approx$ 2 times per minute.}. A more formal definition of problem~\eqref{eq:main_problem} with detailed LP reduction can be found in Appendix~\ref{appendix:lp_optimization}.

After this problem is solved, we assign each peer to aggregate a fraction of gradients proportional to $\min_j g_{ji}$. Peers with $c_i{=}1$ are also tasked with computing the gradients, while peers with $c_i{=}0$ remain idle and only participate in communication. This results in a natural division of labor. In the presence of many compute-heavy peers, some participants without accelerators will dedicate all their bandwidth to gradient aggregation instead of sending their local gradients.

\textbf{Node failures.} The resulting procedure can find the optimal communication strategy for averaging gradients across all participants. However, as the number of participants grows, it might be impractical to compute the global average due to node failures. Based on our experiments with several hundred active volunteers, most training iterations will have at least one participant with network issues. This implies that without necessary precautions, the entire averaging round will fail more often than it will succeed. %This implies that the averaging protocol will fail more often than it will succeed.
To combat this issue, we use techniques~\cite{moshpit,wagma} that replace global averaging with several consecutive iterations in alternating groups of size $m$. The groups are chosen in such a way that the collaboration can obtain the exact average in $\log_m n$ steps. Furthermore, if any single participant fails, it will only affect his immediate group rather than the entire collaboration.

We adaptively choose the optimal group size $m$ based on the number of peers and their failure rates. This optimization problem is independent of Equation~\eqref{eq:main_problem} and aims to maximize the rate at which collaborators can compute the global average. We elaborate on this procedure in Appendix~\ref{appendix:groups}.

\textbf{Comparison with existing techniques.} Our method was designed as a generalization of existing data-parallel strategies that recovers them in special cases. To illustrate this idea, we provide example configurations for which DeDLOC recovers specific well-known strategies:

\begin{enumerate}[leftmargin=*]
    \item \textbf{AR-SGD:} a homogeneous collaboration with reliable peers will use Butterfly All-Reduce~\cite{butterfly_arsgd};
    \item \textbf{Parameter Server:} adding a single participant with a very high bandwidth and low compute performance will turn the previous collaboration into a parameter server~\cite{ps};
    \item \textbf{BytePS:} participants with the same bandwidth as AR-SGD nodes, but without compute accelerators, will behave as auxiliary summation services from BytePS~\cite{byteps};
    \item \textbf{Decentralized SGD:} any collaboration with a sufficiently high failure rate will converge to $m{=}2$. In this mode, all communication is performed between pairs of nodes, similarly to D-PSGD~\cite{dp_sgd}.
\end{enumerate}

\begin{figure}[t]
    \centering
    \includegraphics[height=78px]{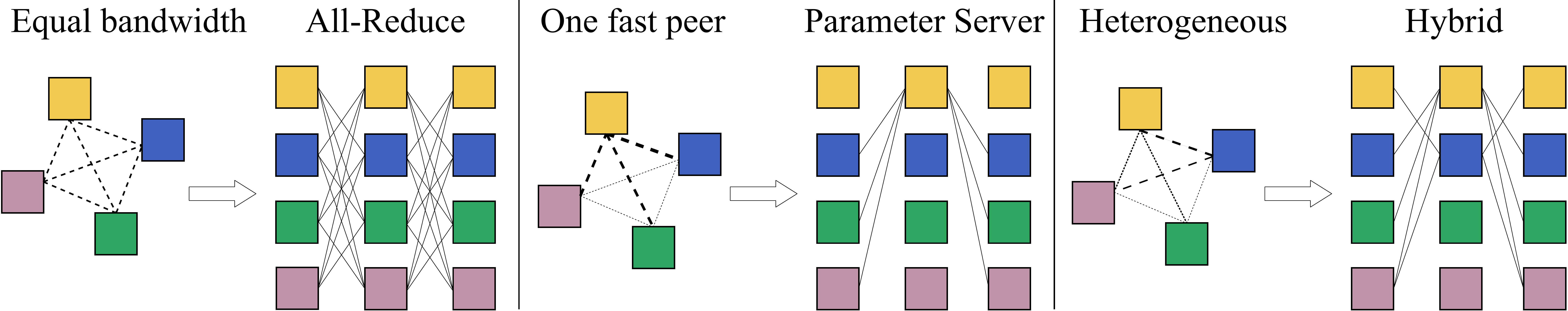}
    \caption{Example collaboration setups and corresponding strategies for optimal averaging. Each square represents one of the peers, line thickness denotes pairwise connection speed.}
    \label{fig:examples}
    \vspace{-12pt}
\end{figure}

However, when training with actual volunteer devices, DeDLOC typically follows a hybrid communication scheme that differs from each of the above options. We display several examples of schemes that can arise as a solution for the optimal strategy search problem in Figure~\ref{fig:examples}.

\vspace{-2pt}
\subsection{System design}\label{sect:method_system_design}

Training with volunteer hardware requires specialized system architecture that can dynamically scale with collaboration size and recover from node failures. DeDLOC achieves these properties by operating as a swarm, similarly in spirit to BitTorrent~\cite{torrent} and I2P~\cite{i2p}. Individual peers coordinate by forming a Distributed Hash Table --- a fully decentralized fault-tolerant key-value storage~\cite{kademlia,kaashoek2003koorde}. Collaborators use this shared ``dictionary'' to count the number of accumulated gradients, find groups for averaging and keep track of the training progress.

DeDLOC ensures that all peers use up-to-date parameters by tracking the number of global steps of each peer. If a peer skips a step, it will observe that others made more steps and download the latest parameters and optimizer statistics from one of the up-to-date peers before resuming training. 
% Similarly, if a peer joins during the experiment, it will download the state from an up-to-date peer.

In order to ensure the integrity of DHT throughout the training run, DeDLOC requires a few peers with stable internet access. These ``backbone'' peers are responsible for welcoming new collaborators and performing auxiliary functions, such as storing checkpoints and tracking learning curves. The only requirement for those peers is that at least one of them is available at all times. As such, the backbone peers can be hosted on inexpensive servers without GPU (see Appendix~\ref{appendix:cost_analysis} for cost analysis).

All other devices are treated as regular collaborators. Depending on their hardware and network bandwidth, these devices can be assigned to (i) compute gradients, (ii) aggregate gradients computed by other peers or (iii) do both, according to the adaptive averaging algorithm. However, performing these steps with actual volunteer devices requires solving another set of challenges described below.

\vspace{-4pt}
\paragraph{Training under NAT and firewalls.}In addition to having uneven compute and network capabilities, volunteer devices also deviate from traditional servers in network configuration. One major difference is the use of Network Address Translation (NAT)~\cite{Biggadike05natblaster:establishing} --- the technology that allows multiple devices to share the same IP address. In practice, the majority of household and organizational computers around the world use one or multiple layers of NAT (see Appendix~\ref{appendix:nat_firewall} for more details). Unfortunately for distributed training, NAT makes it harder to establish peer-to-peer connections~\cite{hole_punching}.

When operating under NAT, DeDLOC participants use one of the following techniques:

\vspace{-4pt}
\begin{enumerate}[leftmargin=*]
    \item \textbf{Hole punching:} use a third peer to temporarily open access to both devices. Once both peers are accessible, they can establish a direct connection and transfer data as usual~\cite{hole_punching};
    \item \textbf{Circuit relays:} both devices connect to a relay (another peer that is mutually accessible), then forward all communication through that relay~\cite{TURN};
    \item \textbf{Client mode:} if everything else fails, a peer can still send gradients to others without the need for incoming connections. This imposes an additional constraint $a_i=0$ for Equation~\eqref{eq:main_problem}.
\end{enumerate}
\vspace{-4pt}

A similar set of strategies can be found in a wide range of distributed systems that rely on peer-to-peer communication, such as WebRTC, VoIP (IP telephony), and BitTorrent. Most of these systems rely on dedicated servers to establish connections between peers. However, in our case it is more appealing to use a fully decentralized NAT traversal where the regular peers perform hole punching and relaying by themselves. We describe this approach in more detail in Appendix~\ref{appendix:p2p}.

\vspace{-4pt}
\paragraph{Training on large datasets.} Many prospective applications of DeDLOC require training on large datasets that can take multiple hours to download. We circumvent this problem by allowing participants to download the data progressively during training. To support this behavior, we split the dataset into shards; upon joining the collaboration, a peer begins downloading examples shard by shard in a streaming fashion. Once the first several examples are obtained, a collaborator can begin training right away while downloading the rest of data in background.

To ensure that the training examples are independent and identically distributed, each participant loads shards in a different random order and uses a buffer to shuffle the data within each shard. Each participant loads the first $S=10,000$ examples into a buffer, then randomly picks a training batch from this buffer and replaces the chosen examples with newly downloaded ones. In our experiments, we stream the training data from a dedicated storage service. However, this service can be replaced with a peer-to-peer data sharing protocol akin to BitTorrent; see Appendix~\ref{appendix:decentralized_dataset_streaming} for details.

% The examples are loaded chunk by chunk from remote files in a streaming fashion. To ensure that we get shuffled datasets, we first make sure that each participant shuffles the dataset shards order. We also use a shuffle buffer: each participant loads the first 10,000 examples into a buffer, then the data loader randomly picks examples from this buffer, and the chosen examples are replaced with new ones in the buffer.

% This allows each participant to start training directly instead of waiting for all the data to be downloaded to their device. 
% Moreover, since the tokenization and masking are done on-the-fly by the participants during training, no data preprocessing is needed before starting the experiment.

% The examples are loaded chunk by chunk from remote files in a streaming fashion. To ensure that we get shuffled datasets, we first make sure that each participant shuffles the dataset shards order. We also use a shuffle buffer: each participant loads the first 10,000 examples into a buffer, then the data loader randomly picks examples from this buffer, and the chosen examples are replaced with new ones in the buffer.

\vspace{-4pt}
\paragraph{Collaborator authentication.} Many prospective applications of DeDLOC need a way to keep track of individual peer contributions and protect against malicious peers. In our experiments, we achieve this using an allowlist authentication system that we describe in Appendix~\ref{appendix:authorization}.

%                                  \\\\\       \\\\\
%      \\\\__.  \\\\__.  \\\\__.  \\\\\\\__.  \\\\\\\__o
% _____\\\\'/___\\\\'/___\\\\'/___\\\\\\\'/___\\\\\\\'/____
% This paper is infested with hedgehogs. Run for your life!

\vspace{-6pt}
\section{Experiments}\label{sect:experiments}
\vspace{-4pt}

In this section, we evaluate the performance of DeDLOC in realistic collaborative training conditions. Our primary focus is on training models that are useful for a wide range of downstream tasks and thus would attract a large number of collaborators. One area that fits this description is self-supervised learning, i.e., learning reusable feature representations on large unlabeled datasets. First, we conduct controlled experiments on two popular self-supervised learning tasks in Sections~\ref{sect:exp_swav} and~\ref{sect:exp_albert}. Then, we set up a real-world collaborative training run with volunteers and report our findings in Section~\ref{sect:exp_bengali}.

\subsection{Self-supervised learning of visual representations}\label{sect:exp_swav}

Our first set of experiments uses SwAV~\cite{swav} --- a self-supervised learning technique that learns image representations by contrasting cluster assignments. Similarly to the original paper, we train the ResNet\nobreakdash-50~\cite{resnet} model on the ImageNet dataset~\cite{imagenet_cvpr09} without labels. Our experiments follow the recommended training configuration~\cite{swav,vissl}: 2+6 random crops, early prototype freezing and a queue with 3,840 samples for each worker, LARS~\cite{lars} optimizer, and 32,768 samples per batch across all workers. In this and further experiments, we use Hivemind~\cite{hivemind} to implement the infrastructure for decentralized averaging. We train with three hardware setups: \textsc{server}, \textsc{workstation} and \textsc{hybrid}. The \textsc{server} setup contains 8 workers, each with a single V100 GPU and 1 Gb/s symmetric bandwidth. In turn, the \textsc{workstation} setup consists of 16 nodes with 1080 Ti and 200 Mb/s bandwidth per worker. Finally, the \textsc{hybrid} setup combines both previous configurations for a total of 24 nodes. Unlike servers, workstation GPUs train in full precision because they do not support accelerated float16 computations~\cite{mixed_precision}.

We report learning curves for each hardware configuration in Figure~\ref{fig:swav_perf}. As expected, the \textsc{hybrid} setup converges the fastest, beating \textsc{server} and \textsc{workstation} setups by 40\% and 52\% accordingly. When used in a supervised setting (Section 4.1 from the original paper), the model learned in this setup achieves a comparable accuracy of 72.2\%.  Another important observation is that the workstation-only experiment achieves reasonable training throughput despite using dated hardware. To provide more insight into the performance of DeDLOC, we also measure the time it takes to run averaging in different configurations. We report the mean over 100 averaging rounds; the standard deviation was below 1\% in all setups. As demonstrated in Table~\ref{tab:averaging_perf}, adaptive averaging does not affect the performance for homogeneous setups while running $1.9$ times faster on the hybrid infrastructure.

\vspace{4pt}
\begin{minipage}[b][][b]{0.5\textwidth}
\centering
\includegraphics[height=100px]{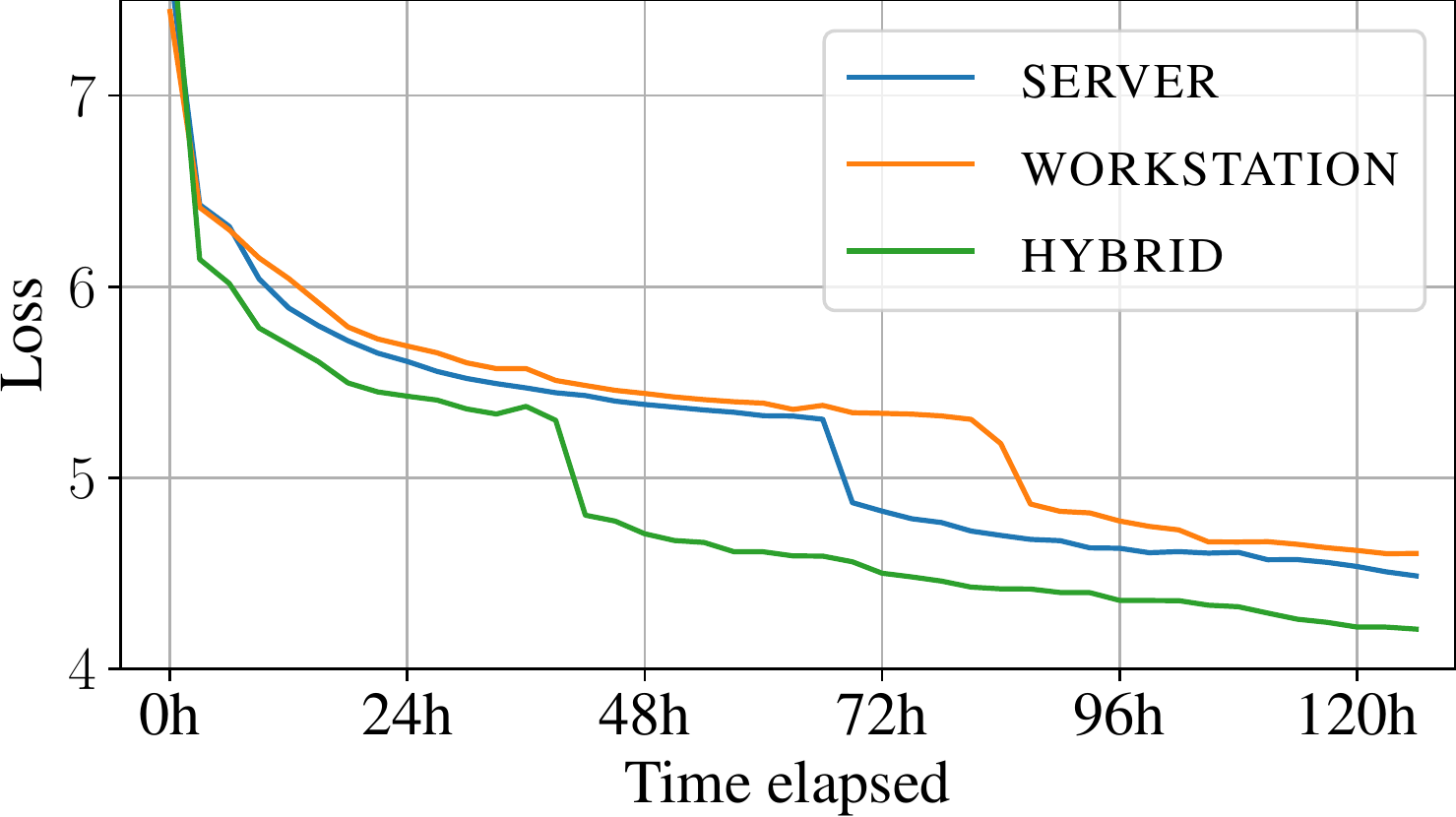}
\captionof{figure}{SwAV pretraining performance.}
\label{fig:swav_perf}
\end{minipage}
\begin{minipage}[b][][b]{0.49\textwidth}
\centering
\renewcommand{\arraystretch}{1.2}
\begin{tabular}{lccc}
\toprule
\multicolumn{1}{c}{\multirow{2}{*}{Setup}} & \multicolumn{3}{c}{Algorithm} \\
 & AR & PS & Ours             \\ 
\midrule
A: 8x1Gb/s & \textbf{1.19} & 4.73 & 1.20 \\
B: 16x0.2Gb/s & \textbf{5.3} & 39.6 & \textbf{5.3} \\
C: A + B& 5.69 & 14.1 & \textbf{2.96} \\
D: B + 1x2.5Gb/s & 5.3 & 3.22 & \textbf{3.18} \\
\bottomrule
\end{tabular}
\vspace{8pt}
\captionof{table}{ResNet-50 averaging performance.}
\label{tab:averaging_perf}
\end{minipage}
\vspace{-4pt}

\subsection{Self-supervised pretraining for language understanding}
\label{sect:exp_albert}

Next, we investigate how collaborative training performs for more complex models. In this experiment, we pretrain the ALBERT-large~\cite{albert} masked language model on the WikiText-103 dataset~\cite{wikitext103}. We chose this setup for two reasons: first, ALBERT is very sensitive to the choice of hyperparameters, and specifically batch size, even more than regular Transformers~\cite{trainingtips}. This makes it easier to verify that DeDLOC can reproduce the training conditions of regular data-parallel training. Second, because of weight sharing, training ALBERT is relatively more compute- and less communication-intensive than regular BERT~\cite{bert}, which makes it possible to train with lower bandwidth.
\nocite{paszke2019pytorch}

As before, we follow the exact training configuration from the original paper, but use GPUs instead of TPUs. We use the implementation of ALBERT from the \texttt{transformers} library~\cite{wolf-etal-2020-transformers}. We run all experiments on cloud instances with Tesla T4 GPUs and report the training loss as a function of time, similarly to~\cite{lin2020multinode,switch}. In order to evaluate how DeDLOC performs with different network speeds, we consider the following setups on the same platform with controlled conditions:
\begin{itemize}[leftmargin=*]
   \item \textbf{High-bandwidth:} 16 workers, each with Tesla T4 and 25 Gb/s symmetric bandwidth;
    \item \textbf{Heterogeneous:} same, but with 4x 200 Mb/s, 8x 100 Mb/s and 4x 50 Mb/s bandwidths;
    \item \textbf{Heterogeneous + load balancing:} like Heterogeneous, but with adaptive averaging (Section~\ref{sect:method_algorithm});
    \item \textbf{Auxiliary peers:} the previous setup with 4 additional CPU-only peers at 1 Gb/s bandwidth.
    \item \textbf{Time-varying:} same as previous, but with 8 additional peers at 100 Mb/s. The extra peers are training part-time, jointly alternating between 8 hours of training and 8 hours of downtime.
\end{itemize}
\pagebreak[0]

\begin{minipage}[b][][b]{0.65\textwidth}
\centering
\includegraphics[width=\textwidth]{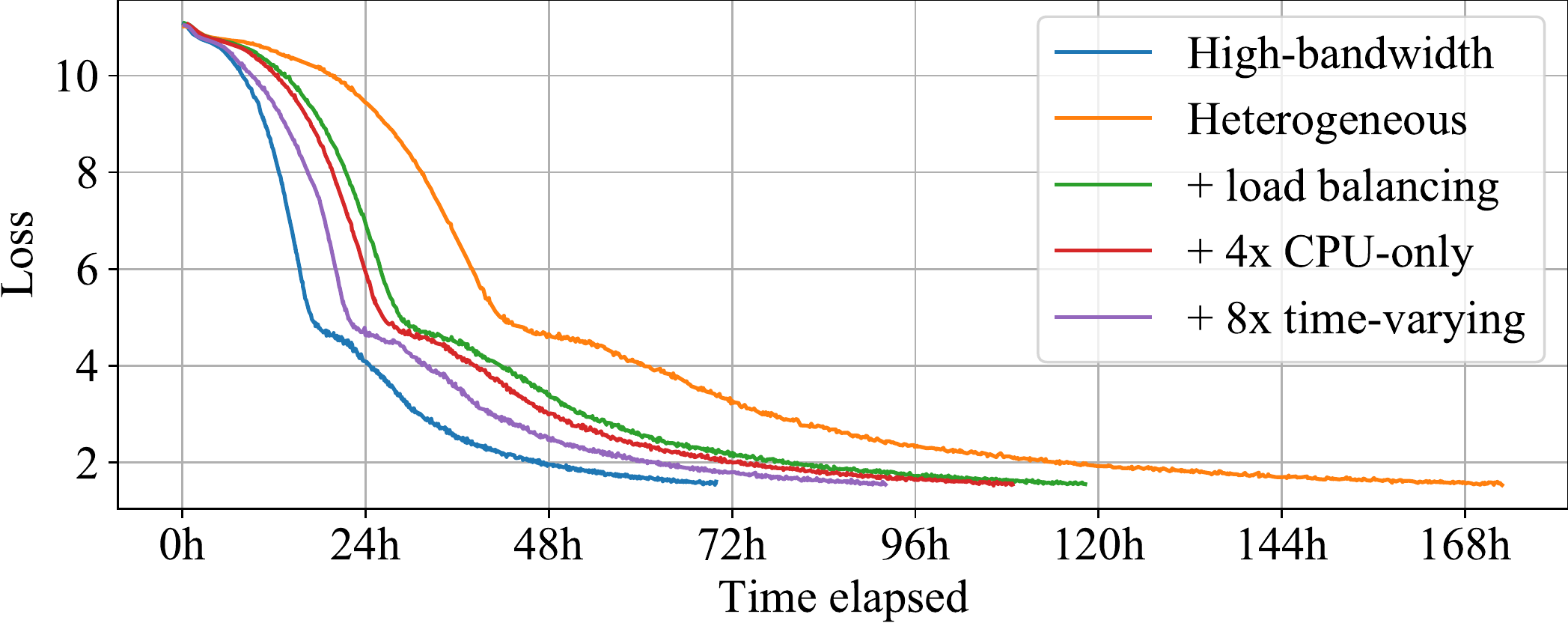}
\vspace{-14pt}
\captionof{figure}{ALBERT pretraining performance.}
\label{fig:albert_perf}
\end{minipage}
\begin{minipage}[b][][b]{0.32\textwidth}
\centering
\includegraphics[width=\textwidth]{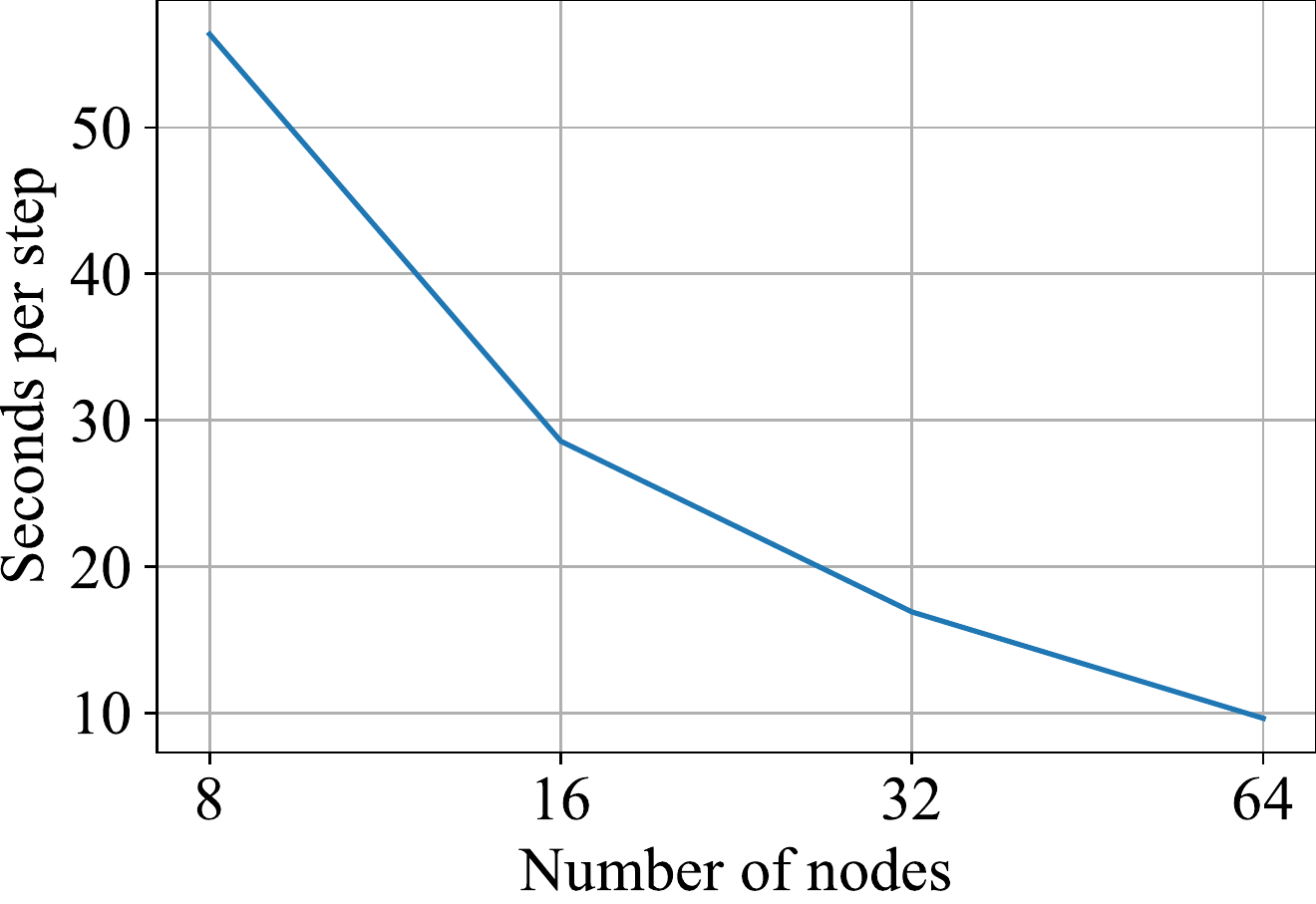}
\vspace{-14pt}
\captionof{figure}{Scalability measurements for ALBERT pretraining.}
\label{fig:albert_scalability}
\end{minipage}

As one can see in Figure~\ref{fig:albert_perf}, naïve training with low-bandwidth peers results in an~$\approx$ 2.5x slowdown compared to high-bandwidth ones. Enabling load balancing accelerates that setup by $\approx47\%$. This effect grows to over 60\% when adding 4 auxiliary peers. Finally, adding 8 part-time peers allows the collaboration to train at 74\% the speed of the high-bandwidth setup without sacrificing the training stability. This turns the latter setup into a viable alternative to traditional distributed training without the need for expensive infrastructure (see the cost analysis in Appendix~\ref{appendix:cost_analysis}). In addition, we demonstrate the high scalability of DeDLOC in Figure~\ref{fig:albert_scalability}, which was obtained by running the same experiment with a varying number of nodes and measuring the time between gradient descent steps. 

\vspace{-6pt}
\subsection{Real-world collaborative training}\label{sect:exp_bengali}
\vspace{-4pt}

For our final evaluation, we organized an actual collaborative training run with volunteer participants, who were asked to pretrain a Transformer masked language model for the Bengali language. This task was chosen deliberately to show the benefits of collaborative training: Bengali has over 230M native speakers who can benefit from recent advances in NLP, but there are few pretrained models available for this language.
We recruited 30 Bengali-speaking volunteers and 10 outside collaborators. All participants received instructions for contributing with free cloud platforms and access to the code for training on local computers. To avoid bias, we did not encourage any specific form of participation: volunteers were free to choose what hardware they contributed and for how long.

Specifically, we trained the ALBERT-large model on Wikipedia and the Bengali part of the OSCAR~\cite{Oscar} multilingual corpus. The model was named sahajBERT after conducting a poll among the participants. We adapted our preprocessing by following the best practices for the Bengali language described in Appendix~\ref{appendix:bn_albert_tokenizer}. To stream from a mix of Wikipedia and OSCAR, the training process iteratively sampled examples from one or the other dataset, as described in Section~\ref{sect:method_system_design}. We accounted for uneven size and quality of data by oversampling Wikipedia by a factor of 2, which resulted in mixing probabilities of 0.23 for Wikipedia and 0.77 for OSCAR. Other hyperparameters were set to the same values as in Section~\ref{sect:exp_albert}. Also, in Appendix~\ref{appendix:xl} we report the results of sahajBERT-XL --- a four times larger model with a specialized architecture that used both GPU and TPU resources.

\begin{figure}[b]
\vspace{-14pt}
% \captionsetup[subfigure]{belowskip=-1pt}
\noindent
\centering
\begin{subfigure}[t]{0.37\textwidth}
\centering
\includegraphics[width=\textwidth]{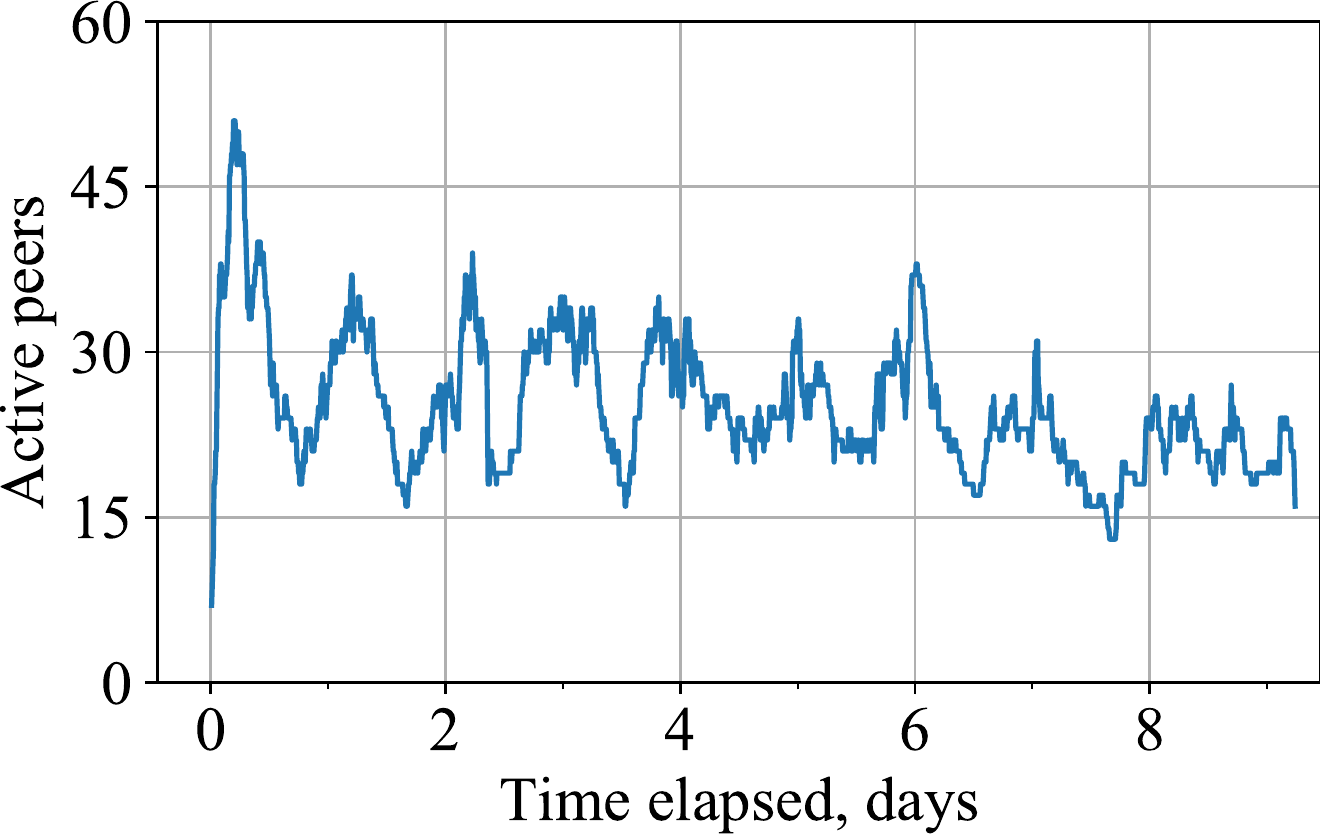}
\caption{Collaboration activity.}
\label{fig:activity}
\end{subfigure}
\begin{subfigure}[t]{0.3\textwidth}
\centering
\includegraphics[width=\textwidth]{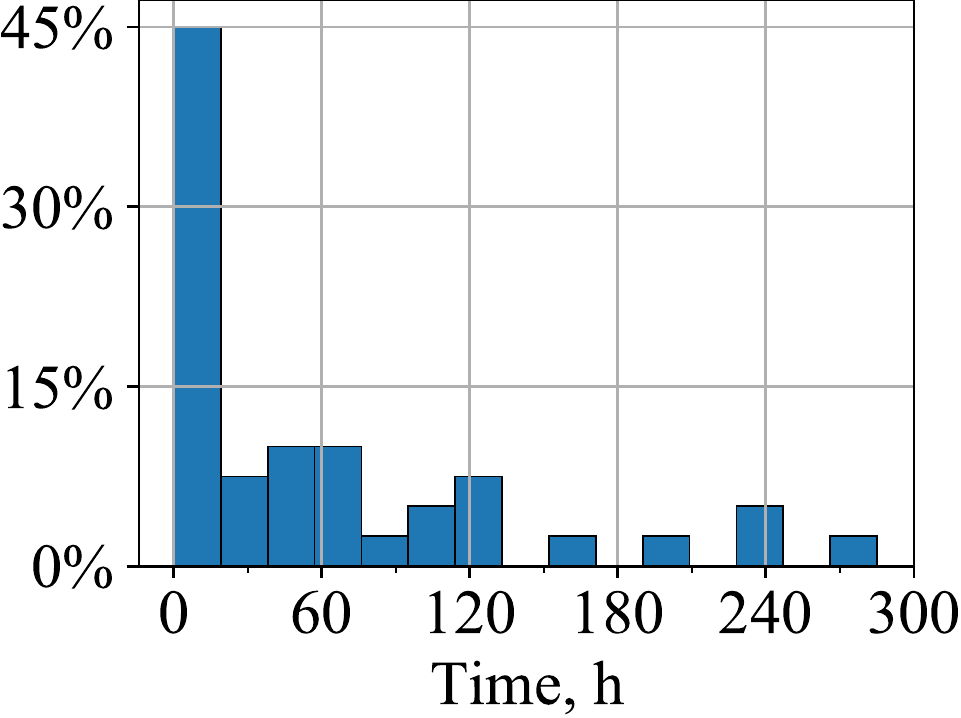}
\caption{Participation time histogram.}
\label{fig:contrib}
\end{subfigure}
\begin{subfigure}[t]{0.3\textwidth}
\centering
\includegraphics[width=\textwidth]{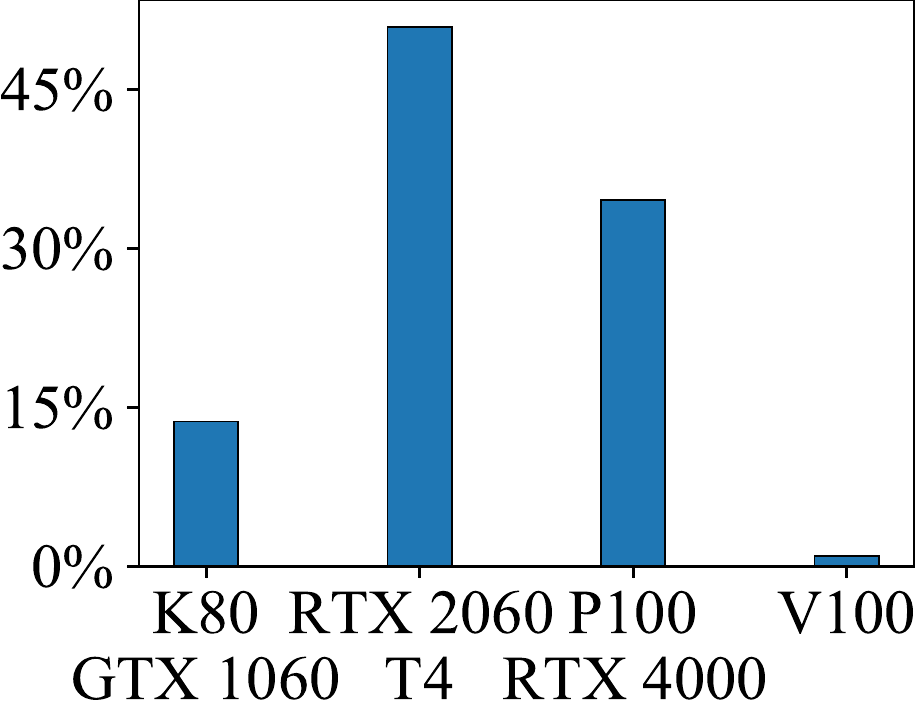}
\caption{Summary of volunteer hardware with example GPUs.}
\label{fig:device}
\end{subfigure}
\vspace{-4pt}
\caption{Collaborative experiment summary.}
\end{figure}

In total, the 40 volunteers contributed compute time from 91 unique devices, most of which were running episodically. Figure~\ref{fig:contrib} shows that although the median GPU time contributed by volunteers across all devices was $\approx$ 1.5 days, some participants ran the training script on several devices, attaining more than 200 hours over the duration of the experiment. With the exception of the start and the end of the collaborative run, the number of simultaneously active devices mostly varied between 15 and 35 depending on the local time. There was less activity in the last 3 days, likely because the volunteers could see that the model has converged on a public Weights \& Biases~\cite{wandb} dashboard.

As depicted in Figure~\ref{fig:device}, individual device performance varied significantly among the collaborators. Along with the resources provided by participants, we also used 16 preemptible single-GPU cloud T4 instances for training.
We have estimated that the average volunteer device consumed 6.95 GB of network traffic per hour of training. While this bandwidth usage is by no means insignificant, it is comparable with cloud gaming~\cite{google_stadia} or high-quality video streaming~\cite{netflix}.

The model converged after 8 days of training, which is 1.8x as fast as regular distributed training with 8 V100 GPUs that we ran as a baseline; Figure~\ref{fig:collab_loss} displays the convergence plots for both setups. At the same time, the stepwise learning curves of the two runs were virtually identical (see Appendix~\ref{appendix:stepwise_learning_curves}), which supports our hypothesis that training with DeDLOC is equivalent to a regular large-batch SGD.
 
\vspace{2pt}
\begin{minipage}[b]{0.49\textwidth}
\centering
\includegraphics[width=0.95\linewidth]{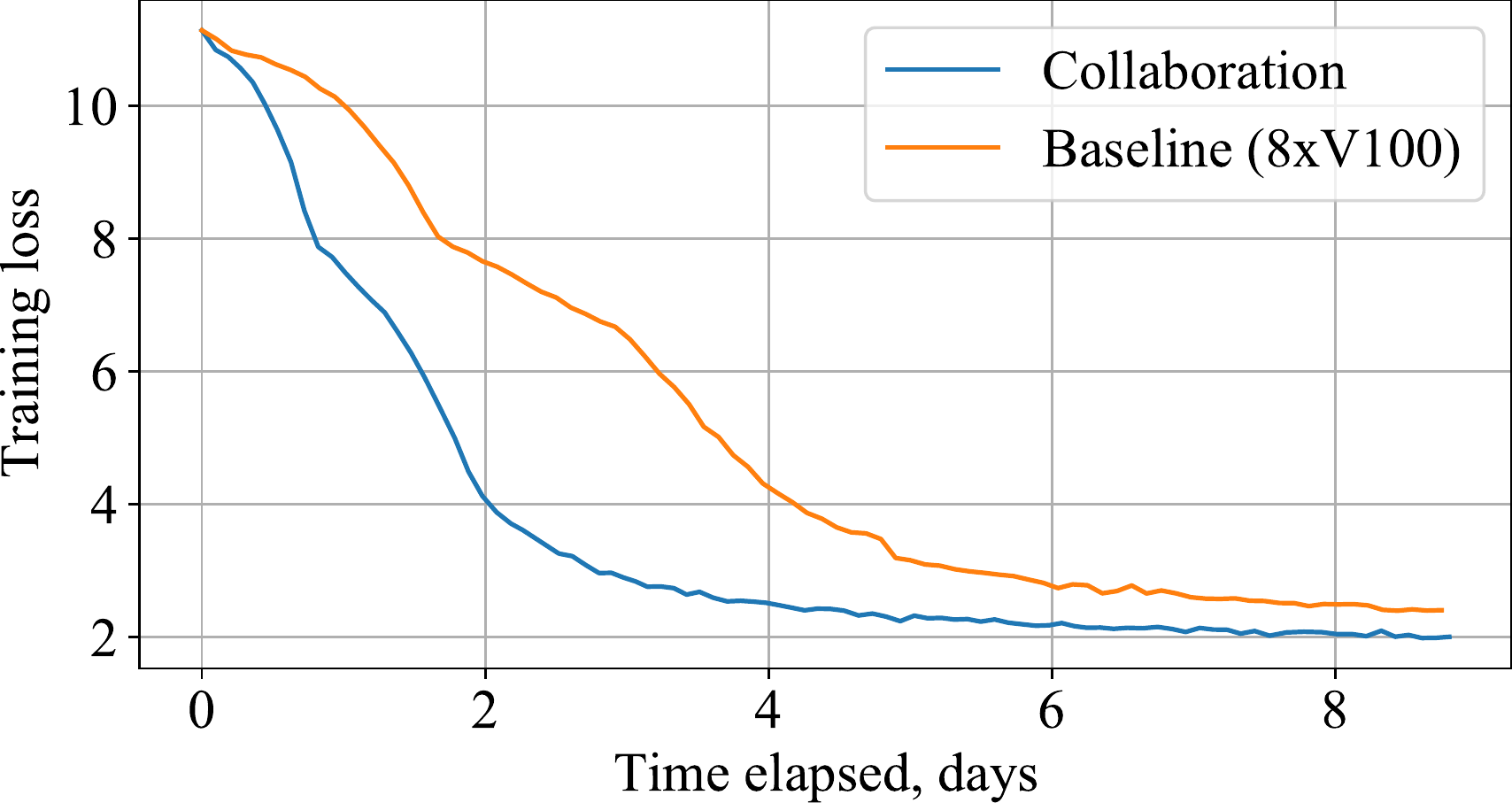}
\vspace{-4pt}
\captionof{figure}{Training progress of sahajBERT.}
\label{fig:collab_loss}
\end{minipage}
\begin{minipage}[b]{0.5\textwidth}
\setlength{\tabcolsep}{2pt}
\renewcommand{\arraystretch}{1.2}
\begin{tabular}{lcc}
\toprule
Model  & Wikiann F1 &NCC Accuracy \\
\midrule
bnRoBERTa           & 82.32 $\pm$ 0.67 &  80.94 $\pm$ 0.45 \\
IndicBERT           & 92.52 $\pm$ 0.45 &  74.46 $\pm$ 1.91 \\
XLM-R               & 96.48 $\pm$ 0.22 &  90.05 $\pm$ 0.38 \\
\midrule
sahajBERT           & 95.45 $\pm$ 0.53 &  91.97 $\pm$ 0.47 \\
sahajBERT-XL        & \bf 96.59 $\pm$ 0.26 & \bf 92.91 $\pm$ 0.43 \\
\bottomrule
\end{tabular}
\captionof{table}{Downstream evaluation results.}
\label{tab:downstream}
\end{minipage}

Finally, we compared the Bengali language representations of sahajBERT with those of other pretrained models on several downstream applications. The first model is XLM-R Large~\cite{xlmr} --- a cross-lingual Transformer-based masked language model that was pretrained on 100 languages and remains a strong baseline for multilingual representation learning. Similarly to sahajBERT, the second model, IndicBERT~\cite{kakwani-etal-2020-indicnlpsuite}, is also based on the ALBERT architecture; however, it was pretrained on 12 languages, including Bengali and Indian English. The third model, bnRoBERTa~\cite{jain2020indictransformers}, is a RoBERTa architecture trained on a monolingual Bengali corpus. We evaluate the model quality on two tasks: WikiANN~\cite{pan-etal-2017-cross} named entity recognition dataset and Soham News Category Classification benchmark from IndicGLUE~\cite{kakwani-etal-2020-indicnlpsuite}. For a detailed description of the setup, refer to Appendix~\ref{appendix:exp_bengali_evaluation}.

As shown in Table~\ref{tab:downstream}, sahajBERT performs comparably to three strong baselines despite being pretrained in a heterogeneous and highly unstable setting.
Notably, our collaboratively trained model outperforms two specialized monolingual baselines and demonstrates competitive results to XLM-R Large, even though the latter has significantly more parameters (560 million instead of 17 million) and was trained on five hundred high-performance data center GPUs instead of tens of low-cost or even free-tier accelerators.
This result confirms previous findings on the benefits of parameter sharing  that were made by authors of ALBERT. Also, it highlights one additional advantage of such architectures: specifically, one can train a high-quality representation model in a communication-constrained setting (for instance, over the Internet) without facing noticeable data transfer bottlenecks.

\section{Conclusion}\label{sect:conclusion}

In this work, we proposed DeDLOC --- a collaborative deep learning approach that enables large-scale collective distributed training on whichever computers available to participants, regardless of hardware and network limitations.
We demonstrated with several experiments that this is a viable approach that maintains its efficiency in a broad range of conditions. Finally, we report the first real collaborative training run of such a scale and share our findings on volunteer activity to pave the road for similar experiments in the future.

An essential property of collaborative training is its environmental impact. While all distributed training experiments have a negative impact due to carbon emissions~\cite{Anthony2020CarbontrackerTA}, DeDLOC has one unique advantage. Due to the ability to utilize heterogeneous low-end devices, it can prolong the effective lifespan of existing computers. We discuss other aspects of environmental impact in Appendix~\ref{appendix:env_impact}.

One issue that needs to be addressed before starting collaborative experiments is the need to gather a community of volunteers. Although our proposed authentication mechanism (see Appendix~\ref{appendix:authorization}) allows acknowledging participants for their contributions (briefly discussed in Appendix~\ref{appendix:contribution_measurement}), the best approach to recruit volunteers is an open question: one needs to take into account both the resources of community members and their motivation for training a specific model.

\section*{Acknowledgements}
We thank Stas Bekman, Dmitry Abulkhanov, Roman Zhytar, Alexander Ploshkin, Vsevolod Plokhotnyuk and Roman Kail for their invaluable help with building the training infrastructure.
Also, we thank Abhishek Thakur for helping with downstream evaluation and Tanmoy Sarkar with Omar Sanseviero, who helped us organize the collaborative experiment and gave regular status updates to the participants over the course of the training run.
Finally, we thank the anonymous reviewers for their feedback on the content and the presentation of our paper.

In addition, authors would like to thank the students of Yandex School of Data Analysis who volunteered to participate in preliminary experiments.

We kindly thank all participants of the Neuropark community\footnote{\href{https://huggingface.co/neuropark}{huggingface.co/neuropark}} who contributed to sahajBERT training. Below, we list the community members who agreed to provide their name for this paper: Aakash Gupta, Aninda Goswamy, Anjali Prasad, Anurag Singh, Arijit Sarkar, Chirranjit Ghosh, Debajit Mallick, Ibraheem Muhammad Moosa, Ishan Bagchi, Khalid Saifullah, Laxya Agarwal, Manan Dey, Mir Ali, Mrinal Mathur, Nilavya Das, Preetha Suri, Priyadarshan Sarkar, Sagnik Roy, Sahil Saha, Sanjeev Kumar, Sanskar Upadhyay, Shyam Sunder Kumar, Soumi Kaibartya, Subhranil Sarkar, Sujit Pal, Syed Modassir Ali, Tanmoy Sarkar, and Vaishali Pal.

Training sahajBERT-XL and hybrid GPU-TPU experiments were made possible by John Kintree, Debajit Mallick, Avijit Saha, Ishan Bagchi, Nilavya Das, Priyadarshan Sarkar, Sagnik Roy, Eduard Pokonechnyy, Arina Ruck. Finally, we would like to acknowledge Tanmoy Sarkar for setting up the backbone peer for sahajBERT-XL on his server and contributing to the evaluation codebase.

The computational resources for internal experiments on cloud instances were provided by the Amazon Research Awards program.

\bibliographystyle{unsrt}
\bibliography{bibliography}

\clearpage
\section*{Contributions}
\label{sec:contributions}

\vspace{-2pt}
\subsection*{Conceptual}

\textbf{Michael Diskin} derived the optimization problem for adaptive averaging.

\textbf{Max Ryabinin} designed and led the research.

\textbf{Thomas Wolf} initially proposed to run collaborative training with the community participants.

\textbf{Gennady Pekhimenko} supervised the work from the systems design point of view.

\vspace{-2pt}
\subsection*{Technical}

\textbf{Alexey Bukhtiyarov} implemented the core large-batch decentralized optimization procedure. 

\textbf{Dmitry Popov} implemented the support of client mode and auxiliary CPU peers for training.

\textbf{Michael Diskin} implemented and conducted the ALBERT pretraining experiments.

\textbf{Anton Sinitsin and Dmitry Pyrkin} implemented and conducted the SwAV pretraining experiments.

\textbf{Quentin Lhoest} designed and implemented the training data streaming logic.

\textbf{Alexander Borzunov and Lucile Saulnier} proposed and implemented the authentication protocol.

\textbf{Max Ryabinin} provided the initial code for cloud-based ALBERT pretraining.

\textbf{Maxim Kashirin, Denis Mazur, and Ilia Kobelev} implemented the libp2p integration.

\textbf{Max Ryabinin} supervised the development of the project and reviewed the code of contributions.

\vspace{-2pt}
\subsection*{sahajBERT}

\textbf{Michael Diskin, Alexey Bukhtiyarov, and Dmitry Popov} created the notebooks with instructions.

\textbf{Lucile Saulnier} built the tokenizer for sahajBERT and implemented Bengali-specific preprocessing.

\textbf{Michael Diskin, Lucile Saulnier, Max Ryabinin, and Alexander Borzunov} managed the running sahajBERT experiment, monitored its performance, answered the questions of participants, and investigated the occurring errors.

\textbf{Albert Villanova del Moral} implemented and conducted downstream finetuning experiments.

\textbf{Michael Diskin} created the dashboards and implemented continuous reporting of experiment metrics.

\textbf{Alexey Bukhtiyarov} added automatic model state fetching and pushing to Model Hub.

\textbf{Yacine Jernite} helped to find the Neuropark community that was interested in collaborative training. 

\vspace{-2pt}
\subsection*{Writing}

\textbf{Max Ryabinin} composed the initial structure of the paper, wrote its abstract and the introduction.

\textbf{Max Ryabinin, Dmitry Popov, and Alexey Bukhtiyarov} discussed the distributed training, volunteer computing, and federated learning aspects of related work, respectively.

\textbf{Max Ryabinin, Lucile Saulnier, and Yacine Jernite} wrote the conclusion of the work.

\textbf{Michael Diskin} discussed the use of group-based All-Reduce for training in larger collaborations.

\textbf{Michael Diskin} conducted the cost analysis of different distributed training approaches.

\textbf{Maxim Kashirin, Denis Mazur, and Ilia Kobelev} described methods for NAT traversal along with peer-to-peer networking.

\textbf{Max Ryabinin, Michael Diskin, and Anton Sinitsin} outlined decentralized data streaming.

\textbf{Yacine Jernite} assessed the environmental implications of DeDLOC.

\textbf{Gennady Pekhimenko and Thomas Wolf} helped improve the general presentation of the work.

\textbf{Max Ryabinin, Michael Diskin, and Gennady Pekhimenko} edited the final version of the paper.
\clearpage

\appendix
\section*{Supplementary Material}

\section{Federated learning}
\label{appendix:related_federated}

Federated learning (FL) is an approach that trains the model on decentralized data stored on many devices without sharing private training data~\cite{FedLearningOriginal}. This scenario is currently gaining more popularity with the rising awareness of data privacy and emerging legal constraints, such as GDPR. Similarly to our setting, FL systems must deal with unreliable heterogeneous hardware. However, their main goal is to ensure the data privacy, which often leads to sacrifices in terms of efficiency.

Most practical FL systems utilize a central parameter server that aggregates local gradients from workers and updates the global model.
As we increase the number of workers, the total system performance becomes bounded by the throughput of this server.
The problem is exacerbated by secure aggregation protocols~\cite{PracticalSecureAggregation, FedLearningDiffPrivacy} that further increase the communication overhead to ensure data privacy.
To account for these limitations, production FL systems perform each update using only a small random subset of peers, while the rest remain idle~\cite{FedLearningAtScale}. Contrary to this, our goal is to maximize the training performance by running computations on all peers.

Another recent line of work explores federated learning algorithms with a decentralized communication topology. %~\cite{TODO}.
Maintaining data privacy in these conditions also requires specialized techniques that introduce communication overhead. For instance, \cite{FedLearningDecentralized} proposes a system where workers cannot share parameters directly, relying on a secure peer-to-peer knowledge distillation instead. %even don't transmit model parameters and perform training via global knowledge distillation.

% In order to maintain security in , these algorithms also adopt privacy protection techniques at a significant cost of utility~\cite{FedLearningAdvancesProblems} 
% The alternative approach is decentralized FL, where no central server is required. These methods should . Some architectures even don't transmit model parameters and perform training via global knowledge distillation~\cite{FedLearningDecentralized}. No practical FL system has successfully used this paradigm and it is still an open research problem.

The above discussion makes it clear that the purpose of the federated learning is orthogonal to ours: we aim to train the global model on publicly available data and achieve the best possible performance.

% \begin{itemize}
%     \item what is it trying to achieve and why
%     \item clearly explain how we are different from federated learning
%     \item they care about data privacy (at a cost of compute/network overhead), we care about training samples/s on public data
%     \item practical FL systems require PS or similar architectures, hard to scale
% \end{itemize}

\section{Optimal averaging strategy via linear programming}
\label{appendix:lp_optimization}
Recall that DeDLOC finds the optimal communication strategy by solving the following problem:

\begin{equation}
\label{eq:main_problem_appendix}
\begin{array}{rclll}
\underset{a, g, c}{\max} & &
                                \min\Bigg(\frac{\sum_{i=1}^n s_i \cdot c_i }{B},\ 
                                     \frac{ \min_i\sum_{j} g_{j i}}{P}\Bigg)&\quad&\\
\textrm{s.t. } &\quad & g_{i j} \leq \min_{k: c_k{=}1} a_{k i}  &\quad&\forall i, j \\
                 & \quad  & \sum_{j \neq i}\left( a_{j i} + g_{j i}\right) \leq d_{i} & \quad&\forall i \\
                 & \quad &   \sum_{j \neq i}\left( a_{i j} + g_{i j}\right) \leq u_{i} & \quad&\forall i \\
                 & \quad &   a_{i j} + g_{i j} \leq t_{i j} & \quad&\forall i, j \\
                 & \quad &   a_{i j} \geq 0 \And g_{i j} \geq 0 \And c_i. \in \{0, 1\}& \quad&\forall i, j \\
\end{array}
\end{equation}
Here, $a_{ij}$ denotes the fraction of network throughput allocated to sending gradients from peer $i$ to peer $j$ for aggregation, $g_{ji}$ is the corresponding fraction for returning the averaged tensors back to sender, and $c_i$ is a binary indicator that represents whether or not peer $i$ computes gradients. The remaining variables are parameters that denote peer compute performance $s_i$, maximum download and upload speeds ($d_i$ and $u_i$ respectively) and regional limitations of peer-to-peer throughput ($t_ij$). Finally, $B$ denotes the global target batch size per step and $P$ is the number of model parameters.

As stated earlier in Section~\ref{sect:method_algorithm}, the DeDLOC peers need to find the optimal strategy during each averaging round. As such, we must ensure that the procedure for solving~\eqref{eq:main_problem_appendix} does not introduce any significant overhead. To that end, we reformulate the problem as a linear program by means of several consecutive reductions, which are described below.

\paragraph{Max-min LP reduction.} First, we replace the original max-min objective with a linear one by following the technique described in~\cite{kaplan1974application}: we maximize a new surrogate variable $\xi$ and replace the inner $\min$ by two additional constraints:

\begin{equation}
\label{eq:main_problem_appendix_minmax}
\begin{array}{rclll}
\underset{a, g, c}{\max} & & \xi &\quad&\\
\textrm{s.t. }  &\quad& \xi \leq \frac{\sum_{i=1}^n s_i \cdot c_i }{B} & & \\
                 &\quad & \xi \leq \frac{ \sum_{j} g_{j i}}{P}  &\quad&\forall i \\
                %  &\quad & g_{i j} \leq \min_{k: c_k{=}1} a_{k i}  &\quad&\forall i, j \\
                %  & \quad  & \sum_{j \neq i}\left( a_{j i} + g_{j i}\right) \leq d_{i} & \quad&\forall i \\
                %  & \quad &   \sum_{j \neq i}\left( a_{i j} + g_{i j}\right) \leq u_{i} & \quad&\forall i \\
                %  & \quad &   a_{i j} + g_{i j} \leq t_{i j} & \quad&\forall i, j \\
                %  & \quad &   a_{i j} \geq 0 \And g_{i j} \geq 0 \And c_i \in \{0, 1\}& \quad&\forall i, j \\
\end{array}
\end{equation}

\paragraph{Binary to LP relaxation.} Second, we must account for the binary variable $c_i$. From a formal perspective, using these indicators transforms our problem into a binary mixed-integer program with a combinatorial worst-case complexity. However, for this specific problem, it is possible to rewrite the constraints in such a way that $c_i$ can be treated as a continuous variable $0\leq c_i\leq 1$:%\nocite{lp_relaxation_largescale}
\begin{equation}
\label{eq:appendix_c_relax}
\forall i, j, k \in 1\dots n \quad g_{ij} \leq a_{ki} + (1 - c_k) \cdot d_i
\end{equation}

For $c_k = 1$, the above equation~\eqref{eq:appendix_c_relax} is exactly equivalent to the original constraint $g_{i j} \leq \min_{k: c_k{=}1} a_{k i}$. In turn, setting $c_k < 1$ for some $k$ effectively removes the corresponding peer $k$ from the $\min$ operator, allowing participant $i$ to aggregate tensors with up to its maximum download speed $d_i$ instead of waiting for peer $k$. The $d_i$ factor in~\eqref{eq:appendix_c_relax} can be replaced with any large positive number as long as the constraint~\eqref{eq:appendix_c_relax} is not saturated for $c_k{=}0$. In practice, $c_k \neq 1$ corresponds to peer $k$ \textbf{not} computing gradients, but still assisting in gradient aggregation.

Applying the two above reductions, we get the following linear program:

\begin{equation}
\label{eq:main_problem_appendix_relaxed}
\begin{array}{rclll}
\underset{a, g, c}{\max} & & \xi &\quad&\\
\textrm{s.t. }  &\quad& \xi \leq \sum_{i=1}^n s_i \cdot c_i\; / \;B & & \\
                 &\quad & \xi \leq \sum_{j} g_{j i}\; / \;P  &\quad&\forall i  \\
                 &\quad & g_{ij} \leq a_{ki} + (1 - c_k) \cdot d_i  &\quad&\forall i, j, k \\
                 & \quad  & \sum_{j \neq i}\left( a_{j i} + g_{j i}\right) \leq d_{i} & \quad&\forall i \\
                 & \quad &   \sum_{j \neq i}\left( a_{i j} + g_{i j}\right) \leq u_{i} & \quad&\forall i \\
                 & \quad &   a_{i j} + g_{i j} \leq t_{i j} & \quad&\forall i, j \\
                 & \quad &   a_{i j} \geq 0& \quad&\forall i, j \\
                 & \quad &   g_{i j} \geq 0& \quad&\forall i, j \\
                 & \quad &  0 \leq c_i \leq 1 & \quad&\forall i \\
\end{array}
\end{equation}

To avoid additional synchronization steps, each peer within DeDLOC solves the above problem~\eqref{eq:main_problem_appendix_relaxed} independently using the interior point solver~\cite{andersen}. Based on the obtained solution, peer $i$ will aggregate a fraction of gradients proportional to its effective throughput: 
\begin{equation}
\text{fraction}_i \propto \frac{\min_j g_{ij}}{ \sum_k \min_j g_{kj}}.
\end{equation}
Furthermore, if $c_i\neq1$, the corresponding participant will disregard its local gradients. In the future, it may be possible to allow such peers to contribute partial gradients akin to~\cite{deepgradientcompression}. However, we leave this investigation to future work.

For certain collaboration compositions, there can be multiple optimal strategies with equal training throughputs. To ensure that all participants act according to the same strategy, we require each peer to solve~\eqref{eq:main_problem_appendix_relaxed} using a deterministic interior point algorithm with globally consistent hyperparameters~\cite{scipy}.

Another practical consideration is that some peers are unable to compute gradients or perform aggregation (for instance, due to networking issues described in Section~\ref{sect:method_system_design}). To account for these limitations, we exclude such peers from aggregation in $\frac{\sum_{i=1}^n s_i \cdot c_i }{B}$ and $\frac{\sum_{j} g_{j i}}{P}$ terms for compute and network resources, respectively.

\section{Fault tolerance}\label{appendix:groups}

In practice, using DeDLOC with large collaborations will eventually require dealing with node failures. If the failures are rare, it is possible to restart the failed steps until they succeed. However, if the collaboration size increases, this strategy will eventually become impractical.

One possible solution is to replace the global (collaboration-wide) All-Reduce with several parallel operations, which is known as Group All-Reduce~\cite{wagma} or Moshpit All-Reduce~\cite{moshpit}. Each operation involves a small independent group of $m$ peers, whereas the groups themselves are formed in such a way that the collaboration can obtain the global average in a logarithmic number of rounds.

Under this strategy, any failed device will only affect its local group instead of the entire collaboration. Furthermore, each individual group will have a higher success rate, since it contains $m\ll n$ peers. In turn, the drawback of using group-based All-Reduce is that the collaboration will need $\lceil\log_m n\rceil$ steps to obtain the global average.

\begin{figure}
    \centering
    \includegraphics[width=0.95\linewidth]{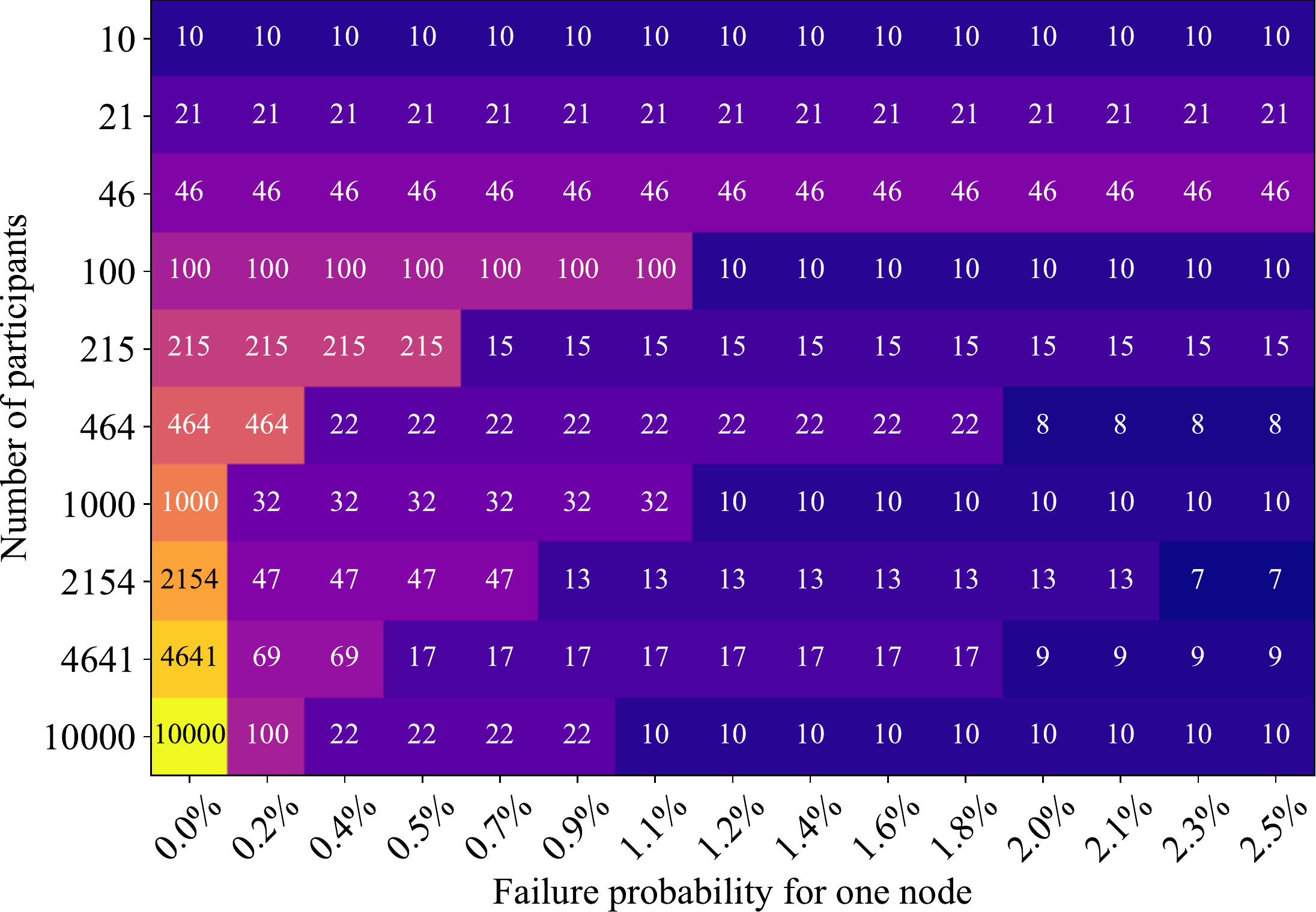}
    \caption{Optimal group size for different collaboration sizes and failure rates.}
    \label{fig:optimal_m}
    \vspace{-16pt}
\end{figure}

We can select the optimal group size by minimizing the \textit{expected} number of iterations required to compute the global average, including both restarts from node failures and the overhead from using Group All-Reduce. For reference, we include the optimal group sizes for typical collaborations and failure rates in Figure~\ref{fig:optimal_m}. In all our experiments, the optimal group size was $m{=}n$ due to a small number of participants and very rare significant network failures.

\section{Network address translation}\label{appendix:nat_firewall}

Collaborative training, similarly to any other application incorporating peer-to-peer communication, is susceptible to a number of networking issues, among which the most common is the inability to accept incoming connections due to Network Address Translation, or NAT~\cite{Biggadike05natblaster:establishing}. The primary function of NAT is to separate the address space of the local network from the global address space by dynamically translating addresses and port numbers of outgoing sessions into public endpoints. Therefore, NAT helps deter the rapid depletion of IPv4 addresses and provides additional security by hiding the local network structure from external parties. However, this also means that NAT devices only authorize outgoing connections, since the dynamic mapping of local endpoints makes it impossible to forward incoming packets to the proper internal host.

For the purposes of the current work, NAT devices can be categorized into two groups --- cone and symmetric. A cone NAT translates an internal IP address and port to the same globally routable endpoint regardless of the destination host, whereas a symmetric NAT allocates different address mapping for each destination host. In case of UDP traffic, the cone NAT can be traversed using the mechanism of UDP Hole Punching. Briefly put, this technique consists of two stages. During the first phase, peers A and B connect to the same globally accessible rendezvous server using the STUN protocol~\cite{STUN}  and exchange their public and private endpoints. The rendezvous server is often called the STUN server by the name of the protocol. At the next step, both peers start sending UDP data packets to each other's endpoints. If A's packet reaches NAT B before B's packet ``punches a hole'', then it is dropped by the NAT B, but when the B's packet reaches NAT A shortly after this, the outgoing session has already been initiated by A, so the B's request is successfully forwarded to A. If both peers happen to ``punch a hole'' in their NATs before the arrival of the counterpart's packet, then the connection is established immediately.

For the TCP traffic, hole punching is also possible, though it has to overcome additional API issues that arise because of the client-server paradigm around which TCP was designed. However, peer-to-peer communication over TCP connections is more robust than over UDP, since NAT usually timeouts the UDP port mapping, thus periodical keep-alive messages must be transmitted. As reported in~\cite{hole_punching}, currently almost two thirds of all NAT vendors provide devices which are compatible with TCP hole punching, that is, consistently map private endpoints and do not send back Reset packets to unsolicited requests.

As for the symmetric NAT, only relaying through a third-party proxy can help establish the connection between peers. This is supported with the TURN protocol~\cite{TURN}. If two peers fail to connect via hole punching, they appeal to the TURN server for an interaction through it.

% Firewall is another obstacle for peer-to-peer networks. It is a system that controls and filters network traffic due to given rules. Each rule can restrict or allow connections. The main goal of a firewall is to protect subnetworks and dedicated hosts from attacks and unauthorized access. 

% Firewalls can be configured with quite strict rulesets. According to the specified rules, connections can be established only by allowlists, or incoming restrictions can be restricted at all. Such settings can lead to the unavailability of the host in peer-to-peer networks.

\section{Peer-to-peer network infrastructure}\label{appendix:p2p}

To enable peer-to-peer interactions that can bypass NAT, we can use the libp2p framework~\cite{libp2p}. Each peer has a set of multiaddresses that allow other participants to establish a connection. Multiaddress comprises an IP address, an L4 protocol (TCP/UDP) with a port, an optional high-level protocol (QUIC), and a peer identifier. A peer can listen to several transport protocols, but it may have only one identifier.

After peers connect to the network, they can interact with each other via their respective identifiers. There are no dedicated STUN and TURN servers in the libp2p network: their role is played by public participants. The network must contain at least 4 publicly accessible peers to be able to recognize public addresses of newly connected peers. Optimally, these are well-known peers with multiaddresses known to all participants. Upon joining, a new node synchronizes with the DHT used for routing and receives information about other available peers. After that, a peer can interact with other participants using their peer id. If the network can get the public address of the peer, then other participants will be able to connect to it.

If a public address of the peer is not available or two peers are using different transport, the communication can be started by relaying requests via an intermediate participant.
Libp2p supports the autorelay feature that allows finding the best relay automatically. When autorelay is enabled, a public peer can serve as a relay for other participants, and a private peer will find the best relay.

\vspace{-2pt}
\section{Cost analysis}\label{appendix:cost_analysis}

In this section, we provide a detailed cost analysis of several hardware and networking setups that can be used for both tasks described in Section~\ref{sect:experiments}, namely, SwAV and ALBERT pretraining.

For simplicity, we only consider temporary resource ownership, i.e., renting GPU-enabled servers instead of building it on-premise. The latter option can be more cost-efficient in the long term, but might be impractical if only a few training runs are required. For the same reason, we do not consider discounts available for committed usage of the same resource over multiple years. As for the rented resources, there are several general hardware categories that we consider:

\begin{enumerate}[leftmargin=*]
    \item High-performance cloud GPU --- dedicated instances with multiple high-end compute accelerators and extremely fast device interconnect.
    \item Low-end cloud GPU --- single-GPU instances with NVIDIA M60, T4 or P40, linked with a fast (preferably intra-datacenter) network of 10--50 Gb/s.
    \item Commodity GPUs --- regular desktop-like machines with consumer-grade GPUs, like NVIDIA RTX 2070, 2080 Ti, 3070. On average, they can have higher performance than low-end cloud devices, but lower network throughput (50--200 Mb/s).
    \item Volunteer hardware --- almost the same class of devices as in the previous section, with the same advantages and disadvantages, but ``free'' for the experiment organizers.
\end{enumerate}

For a fair comparison, we consider three types of GPU instances: cloud V100, cloud T4 and commodity GPUs from peer-to-peer marketplaces, such as \url{vast.ai} or \url{golem.ai}. While several cloud providers offer newer generation GPUs (NVIDIA Ampere), this GPU lineup is still in an active rollout phase, which causes significant price fluctuations. Thus, we base our conclusions on more established generations of GPUs. 

In addition to GPU instances, DeDLOC can also benefit from non-GPU servers that act as auxiliary parameter aggregators. The only real requirement for such servers is high network bandwidth. As such, we consider additional resource types:
\begin{enumerate}[leftmargin=*]
    \item Premium cloud VMs --- low-end instances from premium cloud providers. We consider instances with 2 cores, 16GB RAM and 25 Gb/s maximum bandwidth (symmetric).
    \item Economy cloud VMs --- similar cloud instances (or dedicated servers) from economy cloud providers. For this run, we consider instances with the same 2 cores / 16GB RAM, but only 300--1000 Mb/s symmetric bandwidth (depending on the provider).
    \item Volunteer non-GPU devices --- in theory, it is possible to run collaborative training entirely on volunteer devices with zero hardware expenses for the organizer. However, we omit this option as it trivializes our cost analysis.
\end{enumerate}

On top of that, all cloud and marketplace instances can be rented in a guaranteed (``on-demand'') or a non-guaranteed option. In the latter scenario, the resources are offered at a significant discount, but the resource provider can terminate such instances at any time.

Based on the available resource types and ownership models, we assemble six server fleets with approximately equal training performance in our two experimental setups. For convenience, we order these setups by how difficult they are to operate (easiest-first):

\begin{itemize}[leftmargin=*]
    \item Single high-end node --- 8 x NVIDIA Tesla V100: easiest to operate, but the most expensive option.
    \item Preemptible high-end node has the same hardware but costs less due to irregular availability, which creates a need for regularly saved checkpoints.
    \item Distributed nodes --- 16 x NVIDIA Tesla T4: homogeneous, require distributed optimization.
    \item Distributed + preemptible --- same but preemptible, can be used with a framework that supports elastic training, such as TorchElastic\cite{pytorch_elastic} or Elastic Horovod\cite{elastic_horovod}.
    \item Distributed + heterogeneous --- 5x NVIDIA GTX 1080 Ti, 3x RTX 2070, 1x 2070S, 2x 2080, 4x 2080 Ti, 1x 3070. This configuration has lower bandwidth, thus additional CPU-only peers are needed for efficient averaging.
    \item Collaborative training --- for this setup, we assume that the GPUs from the previous setup are available from volunteers. In that case, the only sources of expenses for the organizer are networking and CPU-only nodes.
    
\end{itemize}

As one can see in Table~\ref{tab:cost}, using a single high-end node is the most expensive alternative. Switching to multiple lower-end nodes and using non-guaranteed instances reduces the cost by a factor of $\approx3$x each. Finally, the volunteer infrastructure is two orders of magnitude cheaper than the high-performance setup. However, some of this price difference is effectively shifted to volunteers. Based on average electricity and networking costs of household Internet connections, we estimate the expense at \$9--30 \textit{per volunteer per month}, assuming 16 volunteers with equivalent GPUs. However, actual costs can vary based on the region, time duration and the exact hardware used by each volunteer.

\begin{table}[b]
\vspace{-14pt}
\renewcommand{\arraystretch}{1.1}
\centering
\caption{Costs of training setups.}
\vspace{6pt}
\label{tab:cost}
\begin{tabular}{@{}lcc@{}}
\toprule
Setup & Instance types & Monthly cost \\ \midrule 
\multirow{2}{*}{Cloud on-demand} & 8xV100 & \$16,898 \\
 & 16xT4 & \$5,299 \\ 
\multirow{2}{*}{Cloud preemptible} & 8xV100 & \$5,133 \\
 & 16xT4 & \$2,074 \\
Marketplace & 4xCPU+16xGPU & \$5,148 \\ 
Volunteer & 4xCPU & \$257 \\ \bottomrule
\end{tabular}
\end{table}

Finally, we want to reiterate that the above setups require different amounts of effort (and expertise). Training on a single high-end node can be done with virtually no code changes in major deep learning frameworks, such as TensorFlow~\cite{tensorflow2015-whitepaper} or PyTorch~\cite{paszke2019pytorch}. In contrast, multi-node (and especially elastic) setups require specialized distributed training frameworks and careful performance tuning. Finally, working with volunteer or marketplace instances introduces a new layer of complexity, which is addressed in this paper.

\paragraph{Networking costs.} When done naïvely, training with geographically distributed participants can incur significant networking expenses. For instance, when using preemptible cloud GPUs from a major provider, allocating these GPUs in different regions can incur additional costs of more than \$3000 per month, compared to a total hardware cost of \$2074 for the same period.

More importantly, using premium non-GPU instances for collaborative training will also incur additional networking costs. Based on our preliminary experiments, a collaborative training setup equivalent to Table~\ref{tab:cost} would lead to an average networking bill of \$5000-6000 per month.
Fortunately, it is possible to circumvent this expense by using cloud providers that do not charge additional costs for network traffic. These providers typically offer less reliable instances with lower maximum bandwidth, which is not a significant issue for DeDLOC.

As a general recipe for reproducing our experiments, we recommend using one of the two setups. When running experiments internally, one can use any major cloud provider as long as all instances are \textit{configured to avoid cross-regional networking costs} (e.g. use internal address space).
In contrast, when training with actual volunteer devices, we recommend using cloud providers without additional networking charges or existing server infrastructure.

\section{Convergence analysis}
\label{appendix:convergence_analysis}
As discussed in Section~\ref{sect:method_general}, DeDLOC updates parameters only after accumulating the gradients for the target number of samples from up-to-date peers. However, due to network-related delays, peers can process more samples than required in some cases. Thus, we can analyze DeDLOC as a regular SGD with varying batch sizes, which allows us to adapt the existing convergence bounds from the optimization literature. More formally, consider a standard optimization problem

\begin{equation}
    \min_{x\in\mathbb{R}^n} f(x),
\end{equation}

which is solved by SGD. We denote the gradients for step $k$ as $\mathbb{E}[g^k|x^k]=\nabla f(x^k)$ and the corresponding update as $x^{k+1}=x^k-\gamma_k g^k$.

Denote the variance of a single stochastic gradient as $\mathbb{E}\left[\left(\nabla f(x^k,\xi_i^k)-\nabla f(x^k)\right)^2|x^k\right]\leq \sigma_0^2$ and the target batch size as $m$. At step $k$, DeDLOC will accumulate gradients from $m_k\geq m$ samples:

\begin{equation}
    g^k=\frac{1}{m_k}\sum_{i=1}^{m_k}\nabla f(x^k,\xi_i^k).
\end{equation}

Thus, the gradient averaged over a minibatch of $m_k$ i.i.d. samples will have the following variance:

\begin{equation}
    \mathbb{E}\left[\left(g^k-\nabla f(x^k)\right)^2| x^k\right]=\frac{1}{m_k^2}\sum_{i=1}^{m_k}\mathbb{E}\left[\left(\nabla f(x^k,\xi_i^k)-\nabla f(x^k)\right)^2| x^k\right]\leq \frac{1}{m_k^2}\sum_{i=1}^{m_k}\sigma_0^2.
\end{equation}

Because $m_k\geq m$,

\begin{equation}
    \frac{1}{m_k^2}\sum_{i=1}^{m_k}\sigma_0^2=\frac{\sigma_0^2}{m_k}\leq\frac{\sigma_0^2}{m},
\end{equation}

which allows us to reuse the existing SGD convergence bounds from the optimization literature~\cite{stich2019unified,khaled2020unified}. For instance, we can use Theorem 5 from~\cite{stich2019unified} and plug in $\frac{\sigma_0^2}{m}$ as gradient variance (with notation also from~\cite{stich2019unified}), getting the following result: 

\begin{equation}
    \mathbb{E}{ f(\bar x_T) - f^\star} + \mu \mathbb{E}{|x_{T+1}-x^\star}|^2 \leq \min \left\{ 64 L R^2 \exp \left[-\frac{\mu T}{4L} \right] + \frac{36 \sigma_0^2}{\mu m T} ,  \frac{2LR^2}{T} + \frac{2 \sigma_0 R}{\sqrt{mT}}  \right\}.
\end{equation}

\section{Decentralized data streaming}
\label{appendix:decentralized_dataset_streaming}
In this section, we propose a generalization of our data streaming approach described in Section~\ref{sect:method_system_design} to a setting without any central data storage. Namely, we offer a way to 
to distribute large datasets across all participants by sharding the examples in the same manner that was used previously.

Specifically, this approach is based on the notion of a local buffer combined with the decentralized metadata storage enabled by the DHT. When a peer joins the experiment, the training process allocates a buffer for several chunks on a local high-capacity storage device (HDD/SSD) available to that peer; the number of chunks is determined by the participant and depends on the hardware capabilities of their computer. Then, in order to procure training data, the peer queries the DHT to find the shards that are stored on the least number of other peers. Assuming that the number of shards does not exceed several thousand, this search can be done by a simple linear-time lookup of all keys without any significant performance drawbacks. After finding such shards, the training process randomly chooses one shard from this set and downloads it from another peer. When the download is complete, the participating node trains on batches from this shard and stores it for later use by other members of the network. The training process repeats such iterations; if the local buffer becomes full at any point, the shards with the highest replication factor are evicted in favor of new data.

The decentralized approach to data streaming has two immediate benefits. First, similarly to distributed training, this approach reduces the load on a single server (or the content delivery network), which might result in significant savings for large-scale experiments that use datasets hosted by cloud providers. Second, even when the data is hosted by organizers of the collaborative experiment, its size might be too large to prevent efficient storage and sharing without investments in specialized infrastructure, which is often quite expensive as well. Storing small portions of the dataset on the computers of participants allows circumventing both issues by distributing the load among all peers.
However, we note that the above approach was not implemented for our current experiments; this section is intended to serve as a description of future work.

\section{Collaborative experiment setup}

\subsection{Instructions for participants}
\label{appendix:volunteer_instruction}
All communication with volunteer contributors took place on a group instant messaging platform. Prior to launching the experiment itself, we used this platform to communicate with Bengali speakers in order to validate the language-specific elements of the model, such as the normalization component of the tokenizer and the sentence splitter tool.

Then, for the collaborative training, we first sent several introductory messages before the event to explain what the event will consist of. Then, we sent a message the day before and a message on the event's launch day with instructions on how to join the training run. Lastly, we sent daily messages to report the current status of the event. The content of the first such message can be found in Figure \ref{figure:instruction_message}. 

In this message, the volunteers were invited to:
\begin{enumerate}[leftmargin=*]
    \item Submit their Hugging Face usernames;
    \item Once added to the allowlist, join the training via notebooks provided by the organizers. After checking that the connection was established and that the GPU was available, participants had to run the notebook and fill in their credentials for the Hugging Face authorization API. 
\end{enumerate}
\begin{figure}[h]
  \centering
  \fbox{\parbox{\textwidth}{Hi @everyone! We’re starting the Collaborative Training Experiment now! Here is some important information:\\

\textbf{How to participate?}\\
1. As a reminder, you need to provide your Hugging Face username to be able to participate. 
For the current participants, @Tanmoy already gathered this list (thank you @Tanmoy!).
For new participants, please join \textit{\#albert-allowlist} and add your username. Someone from the team will add you to the allowlist. If you see a \hourglass~ reaction, we’re on it! If you see a \checkbox, you should be added by then. Feel free to reach out to @Omar Sanseviero, @Mike Diskin, @Quentin Lhoest, @Lucile Saulnier or me if you don’t have access.

2.  You can join the training with:
\begin{itemize}
    \item \textbf{Colab}:  \hl{link}
    \item \textbf{Kaggle}: \hl{link}
    
This option provides you a P100 and lasts longer than Colab. This requires a Kaggle account. You must \textbf{enable Internet access and switch kernel to GPU mode} explicitly.
If it is stuck at ``installing dependencies'' for over 5 minutes, it means you changed the session type too late. Simply restart with GPU/Internet enabled and it should work just fine.
\end{itemize}

Please do not run multiple GPU instances on the same service! You can use Kaggle in one tab and Colab in another, but avoid having two Colab GPU instances at the same time.\\

Local run: if you have a local GPU and you’re tech-savvy. We will keep you informed when this option is available. Stay tuned!\\

Feel free to ask any questions in \textit{\#albert-bengali-training} channel and reach out to us (at the right you can see the members of the Yandex and HF teams).\\
In the following dashboard you can track the status of training: \hl{link}\\

Thank you all for participating and let us know if you have any questions!
}}
%   \fbox{\rule[-.5cm]{0cm}{4cm}  \rule[-.5cm]{4cm}{0cm}}
  \caption{The message sent to participants at the event launch. Parts \hl{in grey} refer to external links.}
  \label{figure:instruction_message}
  \vspace{-12pt}
\end{figure}

\subsection{Measurement of volunteer contributions}
\label{appendix:contribution_measurement}
To let participants follow their own contributions as well as the overall training effort, they were given access to real-time Weights\&Biases dashboards. Each participant could see their personal contributions with the total number of training examples they processed, as well as how much time they contributed and the loss function dynamics of their local models. The volunteers also could compare their contributions: in particular, participants with more computational resources could see the impact they had by comparing the number of samples per second they contributed with other runs. Finally, at the end of the event, a leaderboard of the ones with the highest number of contributed examples was shared with everybody to congratulate the participants.

Although this scheme proved to be highly engaging, it could be improved by also acknowledging the peers that do not contribute the GPU resources but are still very helpful to the collaboration. For example, CPU-only peers with faster network connections can be rewarded for successful averaging rounds and compared between each other in terms of the total number of averaged parameters. Also, to encourage long-term involvement and to increase the stability of the experiment, it might be possible to maintain a list of volunteers with the longest participation time without interruptions.

\subsection{Tokenizer}
\label{appendix:bn_albert_tokenizer}
For this experiment, we used the architecture of the ALBERT model~\cite{albert}; the authors of the original work have chosen the unigram language model \cite{kudo2018subword} token segmentation algorithm that allows transforming a raw text into subword units based on a fixed size vocabulary of 30k tokens.
In order to use the tokenizer that is adapted to the Bengali language, we created a new tokenizer using the Tokenizers library~\cite{hftokenizers2019}. 

This tokenizer is composed of:
\begin{itemize}[leftmargin=*]
    \item Several normalizations adapted to the Bengali language: NMT normalization, NFKC normalization, removal of multiple spaces, homogenization of some recurring unicode characters in the Bengali language and lowercasing;
    \item Specific pre-tokenization rules to condense the vocabulary: we split on whitespaces and replace them with an underscore character ``▁'' (U+2581), we also isolate all punctuation and digits from any other characters;
    \item A Unigram language model as a segmentation algorithm with a 32k tokens vocabulary, trained on the deduplicated Bengali subset of OSCAR~\cite{Oscar};
    \item A template postprocessor, allowing a special token ``[CLS]'' to be included at the start of an example, as well as a special token ``[SEP]'' to separate two segments and to denote the end of sequence.
\end{itemize}

\subsection{Dataset streaming}\label{appendix:dataset_streaming}

Streaming the data to each participant allows them to start training immediately, since the participants do not have to download the full dataset before launching the training. More specifically, the examples from the dataset can be downloaded progressively as training goes. To do so, we used the datasets library~\cite{datasets}. It enabled streaming of Wikipedia and OSCAR, as well as shuffling, on-the-fly processing and mixing of the datasets.

For the experiment, we use the Wikipedia and OSCAR Bengali datasets. Both datasets are split in shards, respectively in the Parquet and GZIP-compressed raw text formats. Information about the datasets is given in Table~\ref{tab:datasets_sizes}. 
The participants download the examples from those files during training, since it is possible to iterate row group by row group from Parquet files and line by line from compressed text files.

The Bengali Wikipedia dataset is based on the 03/20/2021 Wikipedia dump.
The data was processed using the Wikipedia processing script of the datasets library in early April of 2021.
Each example contains the content of one full article, cleaned from markup and sections such as references.

\begin{table}[h]
\vspace{-8pt}
\centering
\caption{Sizes of the Bengali Wikipedia and OSCAR datasets used for training.}
\label{tab:datasets_sizes}
\vspace{6pt}
\begin{tabular}{lcc}
\toprule
\textbf{}               & Wikipedia & OSCAR \\ \midrule
Uncompressed size & 657MB              & 6.2 GB         \\
Documents            & 167,786             & 1,114,481        \\
Shards               & 10                 & 4              \\ \bottomrule
\end{tabular}
\end{table}

To shuffle the datasets, we make each participant iterate over the shards in random order. Then, a shuffle buffer of size $S=10000$ is used, which is compatible with the progressive download of examples. We use a shuffle buffer, because we do not want the participants to download entire shards in the beginning of training just for shuffling.

Sentence splitting, tokenization and preprocessing for next sentence prediction are applied to the examples in an online manner. Since these steps are several orders of magnitude faster than forward and backward passes of the model, they have no significant impact on the training performance.

\subsection{Participant authentication}
\label{appendix:authorization}

Since our experiment was an open collaboration, we chose to set up an authentication system allowing only the people motivated by the final result of the model to join the training. Allowlisting seemed to be the most suitable solution to this need. We therefore distinguish between three types of actors in the distributed network:
\begin{itemize}[leftmargin=*]
    \item \textbf{Central server's moderators}: people who start the experiment, maintain the allowlist and know how to join the training. They have a key pair $(public\_key_{auth}, private\_key_{auth})$ hosted on the central authentication server. In this protocol, the role of the central server is threefold: 1) to verify the identity of a collaborator by requesting the confirmation of an identity provider website, 2) to verify that this collaborator is allowlisted and 3) to distribute access passes to authorized collaborators. Peers have a secure HTTPS-based communication channel with this server in order to protect the data;
    \item \textbf{Digital identity provider}: an entity which is able to create digital identities via a website. In order to create the allowlist, moderators have asked the collaborators to have a digital identity on the identity provider website. This is useful to prevent bots and potential attackers from joining the training and give the moderators opportunity to acknowledge the contribution of each collaborator. In our setup, each identity linked to a username can be claimed by a login and a password owned by one collaborator;
    \item \textbf{Collaborators / Peers}: people who wish to make their computing resources available for collaborative training. Each peer $i$ in the network has a key pair $(public\_key_i, private\_key_i)$. They also have a digital identity on an identity provider website.
\end{itemize}
The following procedures aim to prevent 1) that a non-allowlisted collaborator can interact with the members of the collaborative training and 2) that a malicious actor could claim to be an allowlisted collaborator:

\paragraph{Joining the network:} To join the collaborative training, a peer $i$ must request an access pass from the authorization server. To grant the access pass, the authorization server asks the digital identity provider if the peers are who they claim to be. If the entity provider confirms the peer identity, the authorization server checks that the username appears in the allowlist. If these two steps are verified, the authorization server creates an access pass, otherwise it rejects the peer's request. The access pass is temporary and contains the following information:
\begin{itemize}[leftmargin=*]
    \item The endpoint of a peer already present in the network (a starting point for joining the network);
    \item An access token $access\_token_{i}$ composed of a peer's username, its public key $public\_key_{i}$, and the expiration date of its access pass. The token is signed with the private key $private\_key_{auth}$;
    \item The public key $public\_key_{auth}$.
\end{itemize}
With this access pass, the peer can make requests and respond to them in the decentralized network. After expiration, the peer may repeat this procedure to get a new token.
\paragraph{Making requests:} Alice wants to make a request to Bob. In order for her request to be processed by Bob, we require Alice to include several additional information in her request: 1) her access token $access\_token_{Alice}$, 2) receiver's public key $public\_key_{Bob}$, 3) the current time, 4) a set of random bytes (denoted as \textit{nonce}) that is supposed to be unique for each request and 5) a signature of the request contents and the additional information made with $private\_key_{Alice}$. With this information, Bob considers that a request is not legitimate and should not be processed if one of the following cases occurs:
\begin{itemize}[leftmargin=*]
    \item Alice's access token $access\_token_{Alice}$ is invalid (its signature does not match $public\_key_{auth}$) or expired;
    \item The signature of the request does not match $public\_key_{Alice}$ (stored inside $access\_token_{Alice}$);
    \item The request's current time field differs from the Bob's current time by more than $N$ seconds;
    \item The nonce has already been used during the last $2N$ seconds;
    \item The recipient's public key field does not match the real $public\_key_{Bob}$.
\end{itemize}
These checks protect the exchange against eavesdropped request reuse and man-in-the-middle attacks, because Bob is sure that 1) Alice is specified in the allowlist and her authorization is still valid, 2) the request was created by Alice and could not have been modified by someone else, 3) Bob is the recipient of the request, 4) the request is not repeated by someone who eavesdropped a previous one.
\paragraph{Responding to requests:} When Bob responds to Alice, we also require Bob to include several additional information in his response: 1) his access token $access\_token_{Bob}$, 2) the nonce sent with Alice's request and 3) a signature  of the response contents and the additional information made with $private\_key_{Bob}$. In the same way as above, a response is not considered valid by Alice if:
\begin{itemize}[leftmargin=*]
    \item Bob's access token $access\_token_{Bob}$ is invalid or expired;
    \item The signature of the response does not match $public\_key_{Bob}$ (stored into $access\_token_{Bob}$);
    \item The nonce does not match the nonce stored into Alice's request;
    \item The sender's public key field does not match the real $public\_key_{Bob}$.
\end{itemize}
If the response does not check any of the above cases, Alice is sure that 1) Bob is specified in the allowlist and still has valid access, 2) the response was sent by Bob and could not be modified, and 3) it is the response to the request associated with this nonce. Therefore, an eavesdropped response can't be replayed for another request and a man-in-the-middle attacker can't replace the response content.

\vspace{-4pt}
\subsection{Stepwise learning curves}\label{appendix:stepwise_learning_curves}
\begin{figure}[h]
    \centering
    \includegraphics[height=110px]{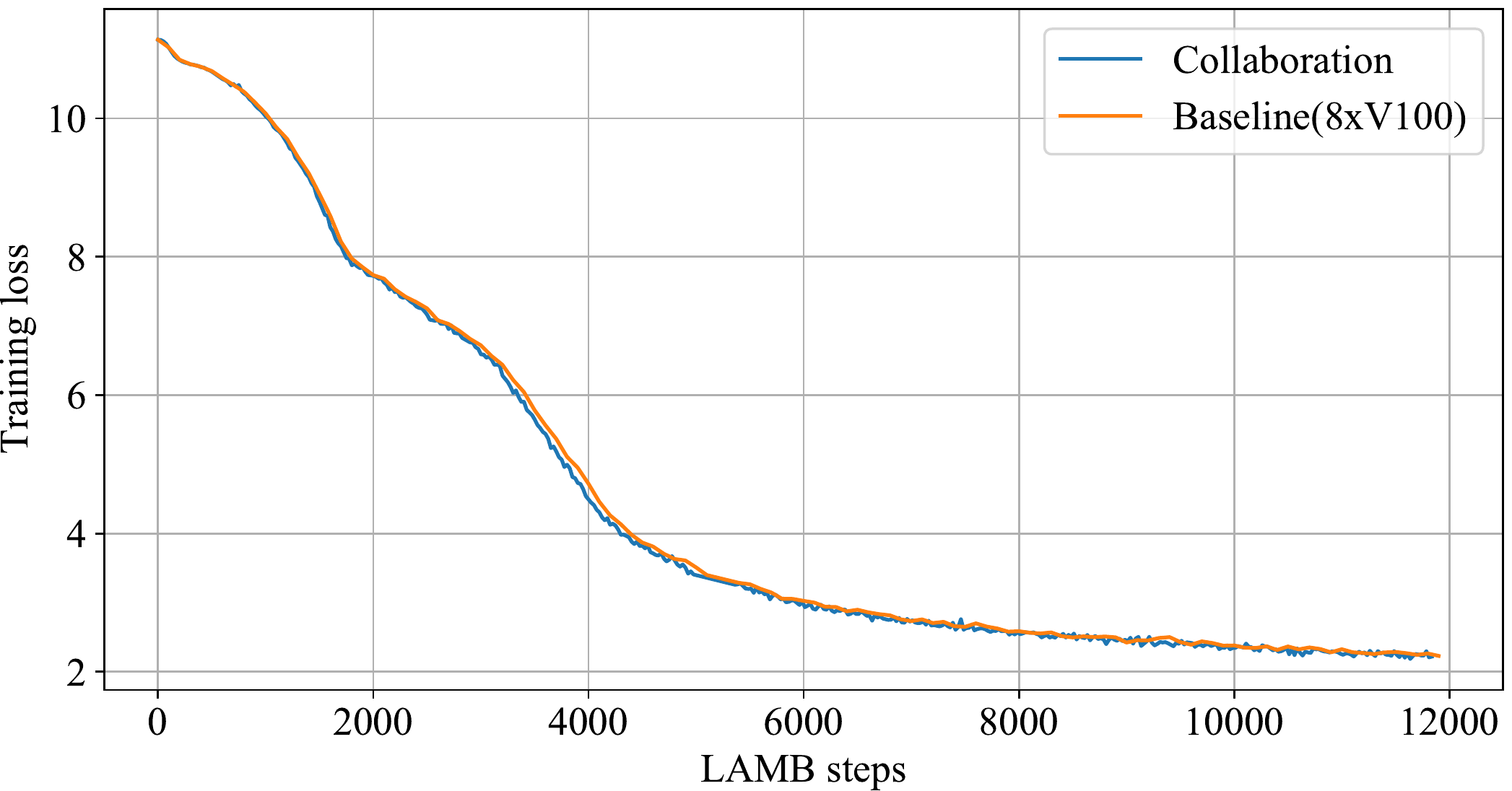}
    \caption{Stepwise training progress of DeDLOC and regular distributed training.}
    \label{fig:stepwise_learning_curves}
    \vspace{-12pt}
\end{figure}

As one can see on Figure~\ref{fig:stepwise_learning_curves}, collaborative training is nearly equivalent to regular data-parallel training in terms of the total number of SGD updates. The slight difference between the two curves is likely due to random variation, though it can also be explained by the fact that DeDLOC uses slightly larger batches due to network latency. In other words, some peers will aggregate a few extra gradients between the moment when the collaboration accumulated 4096 samples and the moment when every peer enters the gradient averaging stage.

\vspace{-4pt}

\subsection{Training sahajBERT-XL with hybrid GPU + TPU resources}
\label{appendix:xl}

To better explore the practical ramifications of collaborative training, we asked volunteers to train a larger model on the same task as the original sahajBERT. We refer to this model as sahajBERT-XL, as it has approximately the same size as ALBERT-xlarge~\cite{albert}: more specifically, the new model has $d_{model}=2048$, $n_{layers}=24$ and three additional architecture modifications:

\vspace{-4pt}
\begin{itemize}[leftmargin=*]
    \item \textbf{Pre-normalization:} the layer normalization~\cite{ba2016layer} was moved to the beginning of each Transformer layer, as in pre-activation residual networks~\cite{he2016identity}. According to prior work, this modification stabilizes the training process in several Transformer applications~\cite{radford2019language,2020t5}.
    \item \textbf{GeGLU activations:} the new model replaces GeLU activation functions with their gated counterparts known as GeGLU, which were shown to improve the performance of Transformer models~\cite{shazeer2020glu,Narang2021DoTM}. However, unlike~\cite{shazeer2020glu}, sahajBERT-XL uses the same number of GeGLU units as in ALBERT-xlarge, which results in 17M additional parameters.
    \item \textbf{Rotary embeddings:} instead of learned absolute positional embeddings, we equip sahajBERT-XL with rotary embeddings~\cite{su2021roformer} that were recently demonstrated to improve training stability of large language models~\cite{rope-eleutherai}.
\end{itemize}
\vspace{-4pt}

The final model had 72.5M parameters, which is $\approx$4 times more than for original sahajBERT. To reduce the computational requirements of sahajBERT-XL pretraining, we initialized it with Net2Net conversion~\cite{net2net} from the original sahajBERT checkpoint after 10,000 training steps. Because of architectural differences, we needed to manually remove the learned positional embeddings and create a new set of GeGLU parameters, which were initialized by copying the existing pre-GeLU parameters and adding Gaussian noise with the variance of $10^{-3}$. 
We increased the training batch size to 16,384 and used the corresponding learning rate schedule from~\cite{lamb}. Before training, we reaccumulated the LAMB optimizer statistics by running 500 steps with a zero learning rate and setting the training schedule to step 3{,}125, which corresponds to the end of the warmup stage.

Despite using this shortcut, training sahajBERT-XL would still require over 3 months using 8 V100 GPUs. To alleviate this problem, we requested volunteers to use both GPU and preemptible TPU (v2 and v3) instances available in several free-tier cloud providers. As a result, a community of 14 volunteers was able to train sahajBERT-XL in 22 days.

However, training in a hybrid GPU-TPU ``cluster'' has proven challenging due to different mixed precision capabilities. Specifically, the available GPU instances could train in float32 and float16 formats, while the TPU cores support float32 and \textbf{b}float16. Unfortunately, training in float16 on GPU and bfloat16 on TPU caused the model to consistently diverge both with Net2Net initialization and when training from scratch: to combat this issue, we switched TPU computations to float32 while keeping GPU ones in float16. Despite this, a TPUv3-8 peer still outperformed any single GPU node.

Using the techniques described above, the volunteers were able to train a model that outperforms both the baselines and the original sahajBERT model on both downstream tasks (see Table~\ref{tab:downstream}). However, due to the significant computational requirements of sahajBERT-XL, we were only able to train the model once without proper hyperparameter sweeps and ablation analysis. Thus, we believe that future research will reveal more efficient strategies for training with hybrid hardware accelerators.

\subsection{Evaluation}\label{appendix:exp_bengali_evaluation}

We compare sahajBERT with three other pretrained language models: XLM-R~\cite{xlmr}, IndicBert~\cite{kakwani-etal-2020-indicnlpsuite}, and bnRoBERTa~\cite{jain2020indictransformers}. For downstream evaluation, we use two tasks from the Indic General
Language Understanding Evaluation (IndicGLUE) benchmark~\cite{kakwani-etal-2020-indicnlpsuite}: named entity recognition (NER) with the balanced train-dev-test splits version~\cite{rahimi-etal-2019-massively} of the original WikiANN dataset~\cite{pan-etal-2017-cross} and news category classification (NCC) with the Soham News Article dataset~\cite{kakwani-etal-2020-indicnlpsuite}.

Each model was finetuned and evaluated as follows:
\begin{enumerate}[leftmargin=*]
    \item For each combination of learning rate in (1e-5, 3e-5) and the maximum input length in (64, 128, 192, 256, 512), we finetuned the model on each task and computed the validation set metrics to find the best hyperparameters. We computed the F1 score for NER and accuracy for NCC;
    \item For the best configuration, we computed the metrics of the corresponding model on the test set. We repeat this step three times for different random seeds, reporting the mean and the standard deviation of metrics.
\end{enumerate}
All finetuning experiments were run using the Adam~\cite{adam} optimizer with the weight decay fix~\cite{loshchilov2017decoupled}, weight decay of 0.001, and the linear decay learning rate schedule. Finally, each model was trained for a maximum number of 20 epochs and stopped earlier if the loss on the validation set did not decrease during 3 epochs. The size of the batch was chosen to be as large as possible: we started with a batch size of 128 and then, if necessary, the batch size is decreased until it can be stored in memory. For the exact hyperparameter values, see Table~\ref{tab:finetuned-model-hyperparams}.

\begin{table}[t]
\vspace{-16pt}
\caption{Hyperparameter values used for model evaluation.}
\label{tab:finetuned-model-hyperparams}
\centering
\vspace{6pt}
\begin{tabular}{llcccc}
\toprule
       Task     & Model  & Learning rate &  Input length & Batch size  \\
 \multirow{5}{*}{NER} 
 & XLM-R & $10^{-5}$ & 256 & 8 \\
 & IndicBERT & $3\cdot 10^{-5}$ & 256 & 64 \\
 & bnRoBERTa & $3\cdot 10^{-5}$ & 512 & 64 \\
 & sahajBERT & $10^{-5}$ & 128 & 32 \\
 & sahajBERT-XL & $3\cdot 10^{-5}$ & 256 & 64 \\
 \multirow{5}{*}{NCC}
 & XLM-R& $10^{-5}$ & 128 & 8 \\
 & IndicBERT & $3\cdot 10^{-5}$ & 128 & 128 \\
 & bnRoBERTa & $3\cdot 10^{-5}$ & 128 & 64 \\
 & sahajBERT & $3\cdot 10^{-5}$ & 64 & 64 \\
 & sahajBERT-XL & $10^{-5}$ & 128 & 64 \\
\bottomrule
\end{tabular}
\end{table}

\section{Environmental impact} 
\label{appendix:env_impact}
Recent works have outlined the environmental consequences of training ever larger deep learning models~\cite{Strubell2019EnergyAP,Schwartz2020GreenA} and encouraged authors to report the incurred energy costs~\cite{Henderson2020TowardsTS}. The direction proposed in this work may help in two specific ways. First, while most of the current tools focus on the CO$_2$ cost caused by the training-time energy consumption~\cite{Anthony2020CarbontrackerTA}, a more holistic evaluation protocol would need to include the not insignificant manufacturing cost of the training infrastructure~\cite{Kline2016HolisticallyET,Bashroush2018ACR}. The collaborative training method described in this work allows volunteers to make better use of existing computing resources, which helps minimize these costs. Second, the distributed training setting allows users to dispense with the extensive cooling infrastructures required for large concentrated data centers, and may thus also help reduce the operating costs themselves~\cite{Qiu2020CanFL}. We note, however, that the additional networking needs may limit the magnitude of these gains.

\end{document}